\documentclass[11pt]{airesearch}

\usepackage{fullpage}
\usepackage{wrapfig}
\usepackage{mathtools}
\setlist[itemize]{leftmargin=*,topsep=1pt,itemsep=1pt}
\setlist[enumerate]{leftmargin=*,topsep=1pt,itemsep=1pt}
\usepackage{xparse}
\RenewDocumentCommand{\paragraph}{s o m}{%
  \par\medskip\noindent\textbf{#3}\quad\ignorespaces
}

\usetikzlibrary{positioning, calc, shapes.geometric, arrows.meta, backgrounds, fit, matrix, decorations.pathreplacing, shadows.blur}

\definecolor{appleBlue}{RGB}{0, 122, 255}
\definecolor{appleLightBlue}{RGB}{235, 245, 255}
\definecolor{appleRed}{RGB}{255, 59, 48}
\definecolor{appleLightRed}{RGB}{255, 235, 235}
\definecolor{appleGreen}{RGB}{52, 199, 89}
\definecolor{appleLightGreen}{RGB}{235, 255, 235}
\definecolor{appleGray}{RGB}{142, 142, 147}
\definecolor{appleLightGray}{RGB}{242, 242, 247}
\definecolor{appleDarkGray}{RGB}{44, 44, 46}
\definecolor{appleDark}{RGB}{28, 28, 30}
\definecolor{applePurple}{RGB}{175, 82, 222}
\definecolor{appleLightPurple}{RGB}{243, 233, 255}
\definecolor{appleOrange}{RGB}{255, 149, 0}
\definecolor{appleLightOrange}{RGB}{255, 235, 205}
\definecolor{appleTeal}{RGB}{48, 176, 199}
\definecolor{appleLightTeal}{RGB}{213, 240, 245}

\usepackage{algorithm}
\usepackage{algpseudocode} 

\ifdefined\final
    \usepackage[disable]{todonotes}
\else
    \usepackage[textsize=tiny]{todonotes}
\fi
\usepackage{fancyhdr}
\newcommand{\projectpage}[1]{%
  \begin{center}
    \small
    \vspace{-1.25ex}
    \texttt{Project Page}: \url{#1}
  \end{center}
}

\title{\huge \bfseries \sffamily Higher-order Linear Attention}
\author{\textbf{Yifan Zhang}$^1$ \quad \textbf{Zhen Qin} \quad \textbf{Mengdi Wang}$^1$ \quad \textbf{Quanquan Gu}$^2$\\[1.5mm]
$^1$Princeton University \quad $^2$University of California, Los Angeles
\\[0.5mm] \texttt{yifzhang@princeton.edu}}

\date{}

\begin{document}
\maketitle


\begin{abstract}
The quadratic cost of scaled dot-product attention is a central obstacle to scaling autoregressive language models to long contexts. Linear-time attention and State Space Models (SSMs) provide scalable alternatives but are typically restricted to first-order or kernel-based approximations, which can limit expressivity. We introduce \textbf{Higher-order Linear Attention} (\textbf{HLA}), a causal, streaming mechanism that realizes higher interactions via compact prefix sufficient statistics. In the second-order case, HLA maintains a constant-size state and computes per-token outputs in linear time without materializing any $n{\times}n$ matrices. We give closed-form streaming identities, a strictly causal masked variant using two additional summaries, and a chunk-parallel training scheme based on associative scans that reproduces the activations of a serial recurrence exactly. We also give the masked third-order streaming kernel and its exact chunk-parallel scan, which uses additional segment maps to compose third-order corrected states. Collectively, these results position HLA as a principled, scalable building block that combines attention-like, data-dependent mixing with the efficiency of modern recurrent architectures.
\end{abstract}

\projectpage{https://github.com/yifanzhang-pro/HLA}

\section{Introduction}

The Transformer architecture~\citep{vaswani2017attention}, powered by scaled dot-product attention, underpins modern large language models (LLMs). Yet the $O(n^2)$ computational and memory complexity in sequence length $n$ constrains long-context use. A rich line of work therefore explores more efficient attention mechanisms (e.g., Linear Attention~\citep{katharopoulos2020transformers, wang2020linformer, choromanski2020rethinking, schlag2021linear, sun2023retentive, qin2023transnormerllm, yang2023gated, qin2024lightning, yang2024parallelizing, von2025mesanet}), Modern Recurrent Neural Networks (RNNs)~\citep{peng2023rwkv, peng2024eagle, sun2024learning,  peng2025rwkv}, Fast Weight Programmers (Delta Networks, \citet{schlag2021linear}), State Space Models (SSMs) \citep{gu2021efficiently, gu2023mamba, dao2024transformers} and Memory Networks~\citep{behrouz2024titans, behrouz2025atlas, behrouz2025s}, which admit $O(1)$ per-token state updates at inference.

We propose \textbf{Higher-order Linear Attention} (\textbf{HLA}), generalizing linear attention by incorporating higher interactions through compact prefix summaries (sufficient statistics). The key observation is that higher attention-like operators admit factorized forms in terms of low-order moments (e.g., sums of key outer products), enabling exact causal streaming without constructing attention matrices. In the second-order case, HLA maintains an constant-size state per head and produces outputs in linear time per token ($O(d^2{+}d\,d_v)$, here $d$ is the query/key dimension and $d_v$ the value dimension, independent to the sequence length).

We address two central challenges: (i) enforcing strict autoregressive causality at second-order without sacrificing streaming updates or introducing any $n{\times}n$ intermediates; and (ii) enabling chunk-parallel training that exactly matches the activations of a serial recurrence. First, we derive an exact masked formulation that enforces strict autoregressive causality by augmenting the state with two additional summaries; the resulting algorithm remains streaming and efficient. Second, we present a chunk-parallel training scheme based on an associative (monoid/semidirect-product) operator that yields the same activations as a serial loop while exploiting intra- and inter-chunk parallelism.

Our contributions are summarized as follows:
\begin{enumerate}
    \item \textbf{Exact masked streaming at second order.} We give a complete algebra of extended summaries that yields strictly causal second-order HLA with per-token constant cost, together with formal statements and proofs establishing masked streaming identities and online updates. The unnormalized HLA is the default operator; the ratio-normalized variant is an option built from the same summaries.
    \item \textbf{Associative scans that match serial activations.} We define an associative (semidirect-product) operator for unmasked and masked settings (with and without exponential decay) and prove that a standard exclusive scan produces forward activations identical to those of a serial recurrence. We also state the reverse-mode algebra.
    \item \textbf{Third-order extension.} We present the full masked third-order state and online updates, a strictly causal streaming kernel, and an exact chunk-parallel algorithm. The third-order scan augments the corrected state with segment-level linear maps; its associative composition reproduces the serial recurrence exactly.
\end{enumerate}

HLA is intended as a drop-in, attention-like mixer for long-context models. It provides (i) attention-style, data-dependent weighting; (ii) strictly causal streaming with $O(1)$ per-token update memory independent of sequence length; and (iii) parallel training via scans without resorting to approximate backpropagation through time. We deliberately focus on algorithmic structure and implementation.

\section{Background}
\label{sec:background}

\textbf{Notations.}\label{sec:notation}
We use bold lowercase for vectors and bold uppercase for matrices/tensors. Token index $t$ denotes the current time; $d$ is the query/key dimension; $d_v$ is the value dimension. Unless otherwise stated, HLA outputs are in the default unnormalized form, which avoids length-dependent renormalization. We adopt row-vector outputs $\mathbf{o}_t\in\RR^{1\times d_v}$; a ratio-normalized (row-normalized) variant divides by a masked scalar denominator built from the same summaries for scale control and comparability with linear attention. Throughout, prefix summaries are statistics computable in streaming fashion with $O(1)$ memory per token and per head.

\subsection{Scaled dot-product attention}
Given queries $\mathbf{Q}\in\RR^{n\times d}$, keys $\mathbf{K}\in\RR^{n\times d}$, and values $\mathbf{V}\in\RR^{n\times d_v}$, scaled dot-product attention~\citep{vaswani2017attention} is
\[
\mathrm{Attn}(\mathbf{Q},\mathbf{K},\mathbf{V})=\mathrm{softmax}\!\left(\frac{\mathbf{Q}\mathbf{K}^\top}{\sqrt{d}} + \boldsymbol{\Lambda}\right)\mathbf{V},
\]
where $\boldsymbol{\Lambda}\in\RR^{n\times n}$ is the additive causal mask (zeros on and below the diagonal; $-\infty$ above). For algebraic manipulations outside the softmax (e.g., Section~\ref{sec:mask}), we use the Hadamard product $\odot$ with a binary causal mask, denoted by $\mathbf{L}$ (ones on and below the diagonal; zeros above), to mask bilinear forms consistently.

\subsection{Linear attention}
Linear attentions~\citep{wang2020linformer, katharopoulos2020transformers, choromanski2020rethinking} approximate the softmax kernel by a feature map $\phi:\mathbb{R}^d\!\to\!\mathbb{R}^r$ (maybe unnormalized):
\[
\mathrm{Attn}(\mathbf{Q},\mathbf{K},\mathbf{V})_i\approx
\frac{\phi(\mathbf{q}_i)^\top \big(\sum_j \phi(\mathbf{k}_j)\mathbf{v}_j^\top\big)}
{\phi(\mathbf{q}_i)^\top \big(\sum_j \phi(\mathbf{k}_j)\big)}.
\]
Maintaining the running sums $\sum_j \phi(\mathbf{k}_j)\mathbf{v}_j^\top$ and $\sum_j \phi(\mathbf{k}_j)$ yields $O(n\,r(d{+}d_v))$ time and $O(r\,d_v)$ memory complexity.

\section{Higher-order Linear Attention}\label{sec:second-order-linear-attention}

%
\begin{figure}[ht!]
\centering
\resizebox{0.79\linewidth}{!}{%
\begin{tikzpicture}[
font=\sffamily,
>=Latex,
base_node/.style={
  thick,
  rounded corners=8pt,
  blur shadow={shadow blur steps=5, shadow opacity=15}
},
tensor/.style={
  base_node,
  draw=appleBlue,
  fill=appleLightBlue,
  minimum height=2.4em,
  minimum width=2.4em,
  font=\sffamily\bfseries
},
value_tensor/.style={
  base_node,
  draw=appleGreen,
  fill=appleLightGreen,
  minimum height=2.4em,
  minimum width=2.4em,
  font=\sffamily\bfseries
},
query_tensor/.style={
  base_node,
  draw=appleRed,
  fill=appleLightRed,
  minimum height=2.4em,
  minimum width=2.4em,
  font=\sffamily\bfseries
},
state/.style={
  base_node,
  draw=appleGray,
  fill=appleLightGray,
  rounded corners=10pt,
  minimum height=3.6em,
  minimum width=3.6em,
  align=center,
  font=\sffamily\bfseries
},
cross_state/.style={
  base_node,
  draw=applePurple,
  fill=appleLightPurple,
  rounded corners=10pt,
  minimum height=3.6em,
  minimum width=3.6em,
  align=center,
  font=\sffamily\bfseries
},
op/.style={
  circle,
  fill=appleLightGray,
  draw=appleGray,
  inner sep=2pt,
  thick,
  font=\sffamily\bfseries\small
},
arrow/.style={
  ->,
  thick,
  color=appleDarkGray,
  line width=1.2pt,
  rounded corners=4pt
},
update_arrow/.style={
  arrow,
  color=appleGreen!65!black
},
cross_arrow/.style={
  arrow,
  color=applePurple,
  dashed,
  line width=1.0pt
},
read_arrow/.style={
  arrow,
  color=appleRed
},
section_label/.style={
  font=\sffamily\bfseries\Large,
  anchor=north west,
  color=appleDark
},
equiv_badge/.style={
  base_node,
  draw=appleDark,
  fill=white,
  rounded corners=14pt,
  inner sep=8pt,
  minimum width=1.6cm,
  minimum height=1.6cm,
  font=\sffamily\bfseries\Huge,
  text=appleDark
},
eqbox/.style={
  base_node,
  fill=white,
  draw=appleGray!50,
  inner sep=9pt
}
]


\begin{scope}[local bounding box=panelA]

\node[font=\sffamily\bfseries\small, text=appleGray] (timeLabelA) at (-1.6,3.0) {time $\rightarrow$};

\node[state] (S0)
  {$\bigl(\mathbf{S}^K_{0},\mathbf{C}^{QV}_{0},$\\$\mathbf{m}^{Q}_{0},\mathbf{G}_{0},\mathbf{h}_{0}\bigr)$};
\node[state, right=2.0cm of S0] (S1)
  {$\bigl(\mathbf{S}^K_{1},\mathbf{C}^{QV}_{1},$\\$\mathbf{m}^{Q}_{1},\mathbf{G}_{1},\mathbf{h}_{1}\bigr)$};
\node[state, right=2.0cm of S1] (S2)
  {$\bigl(\mathbf{S}^K_{2},\mathbf{C}^{QV}_{2},$\\$\mathbf{m}^{Q}_{2},\mathbf{G}_{2},\mathbf{h}_{2}\bigr)$};
\node[font=\sffamily\bfseries, right=0.8cm of S2] (sdots) {$\cdots$};
\node[state, right=0.8cm of sdots] (St)
  {$\bigl(\mathbf{S}^K_{t},\mathbf{C}^{QV}_{t},$\\$\mathbf{m}^{Q}_{t},\mathbf{G}_{t},\mathbf{h}_{t}\bigr)$};

\node[query_tensor, scale=0.8, above=1.1cm of S1, xshift=-14pt] (q1) {$\mathbf{q}_1$};
\node[tensor, scale=0.8, right=0.05cm of q1] (k1) {$\mathbf{k}_1$};
\node[value_tensor, scale=0.8, right=0.05cm of k1] (v1) {$\mathbf{v}_1$};

\node[query_tensor, scale=0.8, above=1.1cm of S2, xshift=-14pt] (q2) {$\mathbf{q}_2$};
\node[tensor, scale=0.8, right=0.05cm of q2] (k2) {$\mathbf{k}_2$};
\node[value_tensor, scale=0.8, right=0.05cm of k2] (v2) {$\mathbf{v}_2$};

\node[query_tensor, scale=0.8, above=1.1cm of St, xshift=-14pt] (qt) {$\mathbf{q}_t$};
\node[tensor, scale=0.8, right=0.05cm of qt] (kt) {$\mathbf{k}_t$};
\node[value_tensor, scale=0.8, right=0.05cm of kt] (vt) {$\mathbf{v}_t$};

\draw[arrow] (S0) -- (S1);
\draw[arrow] (S1) -- (S2);
\draw[arrow] (S2) -- (sdots);
\draw[arrow] (sdots) -- (St);

\draw[update_arrow] ($(q1.south)!0.5!(v1.south)$) -- (S1.north);
\draw[update_arrow] ($(q2.south)!0.5!(v2.south)$) -- (S2.north);
\draw[update_arrow] ($(qt.south)!0.5!(vt.south)$) -- (St.north);

\node[font=\sffamily\bfseries\scriptsize, text=appleGreen!50!black, align=center]
  at ($(S0)+(0.0,-1.20)$) {write outer products\\$+$ cross-term $\mathbf{k}_s(\mathbf{k}_s^{\!\top}\mathbf{C}^{QV}_{s-1})$};

\node[query_tensor, scale=0.9, below=1.2cm of St] (qtR) {$\mathbf{q}_t$};
\node[base_node, dashed, draw=appleGray, fill=white,
      minimum width=1.7cm, minimum height=2.2em,
      right=1.3cm of St] (ytA) {$\mathbf{o}_t$};

\draw[read_arrow] (qtR.north) -- (St.south);
\draw[arrow, color=appleRed!85!black] (St.east) -- (ytA.west);
\node[font=\sffamily\bfseries\scriptsize, text=appleRed,
      above=0.04cm of ytA, anchor=south]
  {read $\mathbf{q}_t^{\!\top}(\mathbf{S}^K_t\mathbf{C}^{QV}_t{-}\mathbf{G}_t)$};

\node[eqbox, below=2.4cm of S2, align=left, xshift=-0.6cm] (eqA) {%
$\displaystyle
\begin{aligned}
\mathbf{S}^K_t &= \mathbf{S}^K_{t-1} + \mathbf{k}_t\mathbf{k}_t^{\!\top},
\quad
\mathbf{C}^{QV}_t = \mathbf{C}^{QV}_{t-1} + \mathbf{q}_t\mathbf{v}_t^{\!\top},
\quad
\mathbf{m}^{Q}_t = \mathbf{m}^{Q}_{t-1} + \mathbf{q}_t \\[2pt]
\mathbf{G}_t &= \mathbf{G}_{t-1} + \mathbf{k}_t\bigl(\mathbf{k}_t^{\!\top}\mathbf{C}^{QV}_{t-1}\bigr),
\quad
\mathbf{h}_t = \mathbf{h}_{t-1} + \mathbf{k}_t\bigl(\mathbf{k}_t^{\!\top}\mathbf{m}^{Q}_{t-1}\bigr) \\[2pt]
\mathbf{o}_t &= \mathbf{q}_t^{\!\top}\!\bigl(\mathbf{S}^K_t \mathbf{C}^{QV}_t - \mathbf{G}_t\bigr)
\end{aligned}$%
};

\node[base_node, fill=appleLightBlue!60, draw=appleBlue!50,
      inner sep=6pt, font=\sffamily\bfseries\scriptsize, text=appleBlue,
      right=0.55cm of eqA, align=center] (complA)
  {$\mathcal{O}(N(d^{2}{+}d\,d_{v}))$ time\\$\mathcal{O}(d^{2}{+}d\,d_{v})$ state at inference};

\end{scope}


\coordinate (xRefEquiv) at ($(panelA.south)+(-3.5,0)$);

\coordinate (yPosAB) at ($(panelA.south)+(0,-2.0)$);
\node[equiv_badge] (equiv) at (xRefEquiv |- yPosAB) {$\equiv$};

\node[eqbox, draw=appleDark!40, fill=appleLightGray!50,
      right=0.45cm of equiv, inner sep=8pt,
      font=\sffamily\small, text=appleDark, align=center] (derivBox)
{\textbf{exact equality for zero-initialised state}:\;
$\displaystyle
\mathbf{S}^K_t=\!\!\sum_{i\le t}\!\mathbf{k}_i\mathbf{k}_i^{\!\top},\;
\mathbf{C}^{QV}_t=\!\!\sum_{i\le t}\!\mathbf{q}_i\mathbf{v}_i^{\!\top},\;
\mathbf{G}_t=\!\!\sum_{i\le t}\!\mathbf{k}_i\mathbf{k}_i^{\!\top}\mathbf{C}^{QV}_{i-1}$\\[2pt]
$\displaystyle
\Longrightarrow\;
\mathbf{o}_t=\sum_{j\le t}\!\Bigl(\sum_{i\le j}(\mathbf{q}_t^{\!\top}\mathbf{k}_i)(\mathbf{q}_j^{\!\top}\mathbf{k}_i)\Bigr)\mathbf{v}_j^{\!\top}
\;=\;\bigl[(\mathbf{L}\odot\mathbf{Q}\mathbf{K}^{\!\top})(\mathbf{L}\odot\mathbf{Q}\mathbf{K}^{\!\top})^{\!\top}\!\odot\mathbf{L}\bigr]_{t,:}\mathbf{V}$%
};


\begin{scope}[shift={(0,-12.2)}, local bounding box=panelB]

\node[base_node, draw=appleRed, fill=appleLightRed,
      minimum width=1.0cm, minimum height=2.4cm,
      align=center, font=\sffamily\bfseries] (Qmat) at (0,0)
  {$\mathbf{Q}$};
\node[font=\sffamily\scriptsize, text=appleRed, above=0.05cm of Qmat]
  {$N\!\times\!d$};

\node[op, right=0.45cm of Qmat] (mul1) {$\times$};

\node[base_node, draw=appleBlue, fill=appleLightBlue,
      minimum width=2.0cm, minimum height=1.0cm,
      align=center, font=\sffamily\bfseries, right=0.45cm of mul1] (KTmat)
  {$\mathbf{K}^{\!\top}$};
\node[font=\sffamily\scriptsize, text=appleBlue, above=0.05cm of KTmat]
  {$d\!\times\!N$};

\node[font=\sffamily\bfseries\Large, text=appleDark, right=0.45cm of KTmat] (eq1) {$=$};

\node[base_node, draw=appleGray, fill=appleLightGray,
      minimum width=2.0cm, minimum height=2.0cm,
      align=center, font=\sffamily\bfseries, right=0.5cm of eq1] (Scores) {};
\begin{scope}
  \clip[rounded corners=8pt] ([xshift=2pt,yshift=-2pt]Scores.north west)
        rectangle ([xshift=-2pt,yshift=2pt]Scores.south east);
  \fill[appleBlue!18]
    ([xshift=2pt,yshift=-2pt]Scores.north west) --
    ([xshift=-2pt,yshift=2pt]Scores.south east) --
    ([xshift=2pt,yshift=2pt]Scores.south west) -- cycle;
\end{scope}
\foreach \i in {0.2,0.5,0.8,1.1,1.4,1.7} {
  \draw[appleRed!22, line width=0.6pt]
    ([xshift=\i cm,yshift=-2pt]Scores.north west) --
    ([xshift=-2pt,yshift=-\i cm]Scores.north east);
}
\node[font=\sffamily\bfseries\scriptsize, text=appleDark, align=center]
  at (Scores.center) {$\mathbf{L}\odot\mathbf{Q}\mathbf{K}^{\!\top}$\\$\eqqcolon\mathbf{W}$};
\node[font=\sffamily\scriptsize, text=appleGray, above=0.05cm of Scores]
  {$N\!\times\!N$};

\node[op, right=0.45cm of Scores] (mul2) {$\times$};

\node[base_node, draw=appleGray, fill=appleLightGray,
      minimum width=2.0cm, minimum height=2.0cm,
      align=center, font=\sffamily\bfseries, right=0.45cm of mul2] (ScoresT) {};
\begin{scope}
  \clip[rounded corners=8pt] ([xshift=2pt,yshift=-2pt]ScoresT.north west)
        rectangle ([xshift=-2pt,yshift=2pt]ScoresT.south east);
  \fill[appleBlue!18]
    ([xshift=2pt,yshift=-2pt]ScoresT.north west) --
    ([xshift=-2pt,yshift=2pt]ScoresT.south east) --
    ([xshift=-2pt,yshift=-2pt]ScoresT.north east) -- cycle;
\end{scope}
\foreach \i in {0.2,0.5,0.8,1.1,1.4,1.7} {
  \draw[appleRed!22, line width=0.6pt]
    ([xshift=2pt,yshift=-\i cm]ScoresT.north west) --
    ([xshift=-\i cm,yshift=2pt]ScoresT.south east);
}
\node[font=\sffamily\bfseries\scriptsize, text=appleDark, align=center]
  at (ScoresT.center) {$\mathbf{W}^{\!\top}$};
\node[font=\sffamily\scriptsize, text=appleGray, above=0.05cm of ScoresT]
  {$N\!\times\!N$};

\node[font=\sffamily\bfseries\Large, text=appleDark, right=0.45cm of ScoresT] (eq2) {$=$};

\node[base_node, draw=applePurple, fill=appleLightPurple,
      minimum width=2.0cm, minimum height=2.0cm,
      align=center, font=\sffamily\bfseries, right=0.45cm of eq2] (T2mat) {};
\begin{scope}
  \clip[rounded corners=8pt] ([xshift=2pt,yshift=-2pt]T2mat.north west)
        rectangle ([xshift=-2pt,yshift=2pt]T2mat.south east);
  \fill[applePurple!22]
    ([xshift=2pt,yshift=-2pt]T2mat.north west) --
    ([xshift=-2pt,yshift=2pt]T2mat.south east) --
    ([xshift=2pt,yshift=2pt]T2mat.south west) -- cycle;
\end{scope}
\node[font=\sffamily\bfseries\scriptsize, text=appleDark, align=center]
  at (T2mat.center) {$\mathbf{T}_{2}{=}\mathbf{W}\mathbf{W}^{\!\top}$\\$\odot\,\mathbf{L}$};
\node[font=\sffamily\scriptsize, text=applePurple!70!black, above=0.05cm of T2mat]
  {$N\!\times\!N$ (causal)};

\node[op, right=0.45cm of T2mat] (mul3) {$\times$};

\node[base_node, draw=appleGreen, fill=appleLightGreen,
      minimum width=1.0cm, minimum height=2.0cm,
      align=center, font=\sffamily\bfseries, right=0.45cm of mul3] (Vmat)
  {$\mathbf{V}$};
\node[font=\sffamily\scriptsize, text=appleGreen!55!black, above=0.05cm of Vmat]
  {$N\!\times\!d_v$};

\node[font=\sffamily\bfseries\Large, text=appleDark, right=0.45cm of Vmat] (eq3) {$=$};

\node[base_node, draw=appleDark, fill=white,
      minimum width=1.0cm, minimum height=2.0cm,
      align=center, font=\sffamily\bfseries, right=0.45cm of eq3] (Out)
  {$\mathbf{O}$};
\node[font=\sffamily\scriptsize, text=appleDark, above=0.05cm of Out]
  {$N\!\times\!d_v$};

\draw[arrow, color=appleRed!75!black] (Qmat.east) -- (mul1.west);
\draw[arrow, color=appleBlue] (KTmat.west) -- (mul1.east);
\draw[arrow] (KTmat.east) -- (eq1.west);
\draw[arrow] (eq1.east) -- (Scores.west);
\draw[arrow] (Scores.east) -- (mul2.west);
\draw[arrow] (mul2.east) -- (ScoresT.west);
\draw[arrow] (ScoresT.east) -- (eq2.west);
\draw[arrow] (eq2.east) -- (T2mat.west);
\draw[arrow] (T2mat.east) -- (mul3.west);
\draw[arrow, color=appleGreen!55!black] (Vmat.west) -- (mul3.east);
\draw[arrow] (Vmat.east) -- (eq3.west);
\draw[arrow] (eq3.east) -- (Out.west);

\node[eqbox, below=1.0cm of T2mat, align=center, xshift=-3cm] (eqB) {%
$\displaystyle
\mathbf{O} \;=\; \bigl[\bigl(\mathbf{L}\odot\mathbf{Q}\mathbf{K}^{\!\top}\bigr)\bigl(\mathbf{L}\odot\mathbf{Q}\mathbf{K}^{\!\top}\bigr)^{\!\top}\odot\,\mathbf{L}\bigr]\,\mathbf{V}
\;\;\equiv\;\;
\mathbf{o}_t = \sum_{j\le t}\sum_{i\le j}(\mathbf{q}_t^{\!\top}\mathbf{k}_i)(\mathbf{q}_j^{\!\top}\mathbf{k}_i)\,\mathbf{v}_j^{\!\top}$%
};

\node[base_node, fill=appleLightRed!55, draw=appleRed!50,
      inner sep=6pt, font=\sffamily\bfseries\scriptsize, text=appleRed,
      right=0.55cm of eqB, align=center] (complB)
  {$\mathcal{O}(N^{2}(d{+}d_{v}))$ time\\fully parallel over $t$};

\end{scope}


\coordinate (yPosBC) at ($(panelB.south)+(0,-2.3)$);
\node[equiv_badge] (equiv2) at (xRefEquiv |- yPosBC) {$\equiv$};

\node[eqbox, draw=appleDark!40, fill=appleLightGray!50,
      right=0.45cm of equiv2, inner sep=9pt,
      font=\sffamily\small, text=appleDark, align=center] (derivBox2)
{\textbf{regroup tokens into }$M$\textbf{ chunks of size }$C$,\; $N=MC$,\\[3pt]
\textbf{associative scan with semidirect product over chunks}:\\[3pt]
$\displaystyle
\begin{aligned}
(\cdot)_{A}\,\oplus\,(\cdot)_{B}\;=\;\Bigl(&\mathbf{S}^K_{A}{+}\mathbf{S}^K_{B},\;\;
 \mathbf{C}^{QV}_{A}{+}\mathbf{C}^{QV}_{B},\;\;
 \mathbf{m}^{Q}_{A}{+}\mathbf{m}^{Q}_{B},\\
 &\mathbf{G}_{A}{+}\mathbf{G}_{B}+\,\boxed{\mathbf{S}^K_{B}\mathbf{C}^{QV}_{A}}\,,\;\;
 \mathbf{h}_{A}{+}\mathbf{h}_{B}+\,\boxed{\mathbf{S}^K_{B}\mathbf{m}^{Q}_{A}}\Bigr)
\end{aligned}$%
};


\begin{scope}[shift={(0,-23.2)}, local bounding box=panelC,
  chunk_block/.style={
    base_node,
    draw=appleGray!60,
    fill=appleLightGray!45,
    minimum width=2.8cm,
    minimum height=2.55cm,
    rounded corners=10pt
  }
]

\node[state, minimum width=2.4em, minimum height=2.4em] (Sc0) at (0,0)
  {$\mathcal{S}^{(0)}$};

\node[chunk_block, right=0.8cm of Sc0] (Ck1) {};
\node[font=\sffamily\bfseries\footnotesize, text=appleDark]
  at ([yshift=0.75cm]Ck1.center) {Chunk 1};
\node[base_node, draw=appleRed, fill=appleLightRed,
      minimum width=0.32cm, minimum height=0.70cm]
  at ([xshift=-0.86cm, yshift=-0.0cm]Ck1.center) (Q1m) {};
\node[font=\sffamily\bfseries\scriptsize, text=appleRed,
      below=0.1cm of Q1m] {$\mathbf{Q}^{(1)}$};
\node[base_node, draw=appleBlue, fill=appleLightBlue,
      minimum width=0.70cm, minimum height=0.32cm]
  at ([xshift=0.00cm, yshift=-0.0cm]Ck1.center) (K1m) {};
\node[font=\sffamily\bfseries\scriptsize, text=appleBlue,
      below=0.1cm of K1m] {$\mathbf{K}^{(1)\!\top}$};
\node[base_node, draw=appleGreen, fill=appleLightGreen,
      minimum width=0.32cm, minimum height=0.70cm]
  at ([xshift=0.86cm, yshift=-0.0cm]Ck1.center) (V1m) {};
\node[font=\sffamily\bfseries\scriptsize, text=appleGreen!55!black,
      below=0.1cm of V1m] {$\mathbf{V}^{(1)}$};

\node[state, minimum width=2.4em, minimum height=2.4em, right=0.8cm of Ck1] (Sc1)
  {$\mathcal{S}^{(1)}$};

\node[chunk_block, right=0.8cm of Sc1] (Ck2) {};
\node[font=\sffamily\bfseries\footnotesize, text=appleDark]
  at ([yshift=0.75cm]Ck2.center) {Chunk 2};
\node[base_node, draw=appleRed, fill=appleLightRed,
      minimum width=0.32cm, minimum height=0.70cm]
  at ([xshift=-0.86cm, yshift=-0.0cm]Ck2.center) (Q2m) {};
\node[font=\sffamily\bfseries\scriptsize, text=appleRed,
      below=0.1cm of Q2m] {$\mathbf{Q}^{(2)}$};
\node[base_node, draw=appleBlue, fill=appleLightBlue,
      minimum width=0.70cm, minimum height=0.32cm]
  at ([xshift=0.00cm, yshift=-0.0cm]Ck2.center) (K2m) {};
\node[font=\sffamily\bfseries\scriptsize, text=appleBlue,
      below=0.1cm of K2m] {$\mathbf{K}^{(2)\!\top}$};
\node[base_node, draw=appleGreen, fill=appleLightGreen,
      minimum width=0.32cm, minimum height=0.70cm]
  at ([xshift=0.86cm, yshift=-0.0cm]Ck2.center) (V2m) {};
\node[font=\sffamily\bfseries\scriptsize, text=appleGreen!55!black,
      below=0.1cm of V2m] {$\mathbf{V}^{(2)}$};

\node[state, minimum width=2.4em, minimum height=2.4em, right=0.8cm of Ck2] (Sc2)
  {$\mathcal{S}^{(2)}$};

\node[font=\sffamily\bfseries, right=0.5cm of Sc2] (cdotsC) {$\cdots$};

\node[chunk_block, right=0.5cm of cdotsC] (CkM) {};
\node[font=\sffamily\bfseries\footnotesize, text=appleDark]
  at ([yshift=0.75cm]CkM.center) {Chunk $M$};
\node[base_node, draw=appleRed, fill=appleLightRed,
      minimum width=0.32cm, minimum height=0.70cm]
  at ([xshift=-0.86cm, yshift=-0.0cm]CkM.center) (QMm) {};
\node[font=\sffamily\bfseries\scriptsize, text=appleRed,
      below=0.1cm of QMm] {$\mathbf{Q}^{(M)}$};
\node[base_node, draw=appleBlue, fill=appleLightBlue,
      minimum width=0.70cm, minimum height=0.32cm]
  at ([xshift=0.00cm, yshift=-0.0cm]CkM.center) (KMm) {};
\node[font=\sffamily\bfseries\scriptsize, text=appleBlue,
      below=0.1cm of KMm] {$\mathbf{K}^{(M)\!\top}$};
\node[base_node, draw=appleGreen, fill=appleLightGreen,
      minimum width=0.32cm, minimum height=0.70cm]
  at ([xshift=0.86cm, yshift=-0.0cm]CkM.center) (VMm) {};
\node[font=\sffamily\bfseries\scriptsize, text=appleGreen!55!black,
      below=0.1cm of VMm] {$\mathbf{V}^{(M)}$};

\node[state, minimum width=2.4em, minimum height=2.4em, right=0.8cm of CkM] (ScM)
  {$\mathcal{S}^{(M)}$};

\draw[arrow, update_arrow] (Sc0) -- (Ck1);
\draw[arrow, update_arrow] (Ck1) -- (Sc1);
\draw[arrow, update_arrow] (Sc1) -- (Ck2);
\draw[arrow, update_arrow] (Ck2) -- (Sc2);
\draw[arrow, update_arrow] (Sc2) -- (cdotsC);
\draw[arrow, update_arrow] (cdotsC) -- (CkM);
\draw[arrow, update_arrow] (CkM) -- (ScM);

\draw[cross_arrow] (Sc1.north) to[bend left=35]
  node[midway, above=2pt, font=\sffamily\bfseries\scriptsize, text=applePurple]
    {cross-term $\mathbf{S}^K_{B}\,\mathbf{C}^{QV}_{A}$ (and $\mathbf{S}^K_{B}\,\mathbf{m}^{Q}_{A}$)} (Sc2.north);

\node[base_node, dashed, draw=appleGray, fill=white,
      minimum width=1.4cm, minimum height=1.9em,
      below=1.1cm of Ck1, font=\sffamily\bfseries] (O1c) {$\mathbf{O}^{(1)}$};
\node[base_node, dashed, draw=appleGray, fill=white,
      minimum width=1.4cm, minimum height=1.9em,
      below=1.1cm of Ck2, font=\sffamily\bfseries] (O2c) {$\mathbf{O}^{(2)}$};
\node[base_node, dashed, draw=appleGray, fill=white,
      minimum width=1.4cm, minimum height=1.9em,
      below=1.1cm of CkM, font=\sffamily\bfseries] (OMc) {$\mathbf{O}^{(M)}$};

\draw[arrow, color=appleRed!85!black] (Ck1.south) -- (O1c.north);
\draw[arrow, color=appleRed!85!black] (Ck2.south) -- (O2c.north);
\draw[arrow, color=appleRed!85!black] (CkM.south) -- (OMc.north);

\node[eqbox, below=3.0cm of Sc2, align=center, xshift=-3.5cm] (eqC) {%
$\displaystyle
\begin{aligned}
\mathcal{S}^{(k)} &= \mathcal{S}^{(k-1)} \,\oplus\, \mathcal{T}^{(k)}
\qquad\text{(state tuple } \mathcal{S}=(\mathbf{S}^K,\mathbf{C}^{QV},\mathbf{m}^{Q},\mathbf{G},\mathbf{h})\text{)} \\[2pt]
\mathbf{O}^{(k)}_{t} &= \underbrace{\mathbf{q}_{t}^{\!\top}\!\bigl(\mathbf{S}^{K,(k-1)}\mathbf{C}^{QV,(k-1)} - \mathbf{G}^{(k-1)}\bigr)}_{\text{cross-chunk (carry from }\mathcal{S}^{(k-1)}\text{)}}
\;+\;
\underbrace{\bigl[(\mathbf{L}\odot\mathbf{Q}^{(k)}\mathbf{K}^{(k)\!\top})(\mathbf{L}\odot\mathbf{Q}^{(k)}\mathbf{K}^{(k)\!\top})^{\!\top}\!\odot\mathbf{L}\bigr]_{t,:}\mathbf{V}^{(k)}}_{\text{intra-chunk (parallel masked 2nd-order)}}
\end{aligned}$%
};

\node[base_node, fill=appleLightGreen!55, draw=appleGreen!55!black,
      inner sep=6pt, font=\sffamily\bfseries\scriptsize, text=appleGreen!45!black,
      right=0.55cm of eqC, align=center] (complC)
  {$\mathcal{O}\!\bigl(NC(d{+}d_{v}){+}M(d^{2}{+}d\,d_{v})\bigr)$ time\\$M=N/C$ sequential chunk steps};

\end{scope}


\path
  let \p1=(panelA.west), \p2=(panelB.west), \p5=(panelC.west),
      \p3=(panelA.east), \p4=(panelB.east), \p6=(panelC.east)
  in
    coordinate (commonW) at ({min(\x1,min(\x2,\x5))-0.1},0)
    coordinate (commonE) at ({max(\x3,max(\x4,\x6))+0.1},0);

\node[inner sep=0pt, minimum size=0pt] (anchorA_L) at (commonW |- panelA.center) {};
\node[inner sep=0pt, minimum size=0pt] (anchorA_R) at (commonE |- panelA.center) {};
\node[inner sep=0pt, minimum size=0pt] (anchorB_L) at (commonW |- panelB.center) {};
\node[inner sep=0pt, minimum size=0pt] (anchorB_R) at (commonE |- panelB.center) {};
\node[inner sep=0pt, minimum size=0pt] (anchorC_L) at (commonW |- panelC.center) {};
\node[inner sep=0pt, minimum size=0pt] (anchorC_R) at (commonE |- panelC.center) {};

\begin{scope}[on background layer]
  \node[draw=appleGray!30, left color=appleLightGray!30, right color=appleLightGray,
        rounded corners=16pt,
        fit=(panelA)(eqA)(complA)(anchorA_L)(anchorA_R),
        inner sep=18pt] (boxA) {};
  \node[draw=appleGray!30, left color=appleLightGray!30, right color=appleLightGray,
        rounded corners=16pt,
        fit=(panelB)(eqB)(complB)(anchorB_L)(anchorB_R),
        inner sep=18pt] (boxB) {};
  \node[draw=appleGray!30, left color=appleLightGray!30, right color=appleLightGray,
        rounded corners=16pt,
        fit=(panelC)(eqC)(complC)(anchorC_L)(anchorC_R),
        inner sep=18pt] (boxC) {};
\end{scope}

\node[section_label] at ([yshift=18pt, xshift=10pt]boxA.north west)
  {A. Recurrent Form (causal HLA scan)};
\node[section_label] at ([yshift=18pt, xshift=10pt]boxB.north west)
  {B. Parallel Form (masked 2nd-order tensor attention)};
\node[section_label] at ([yshift=18pt, xshift=10pt]boxC.north west)
  {C. Chunk-wise Parallel Form (associative scan w/ semidirect product)};

\end{tikzpicture}
}
\caption{\textbf{Three equivalent views of Higher-order Linear Attention (HLA, second order).}
(A) The recurrent form maintains a constant-size state tuple
$\mathcal{S}_t=(\mathbf{S}^K_t,\mathbf{C}^{QV}_t,\mathbf{m}^{Q}_t,\mathbf{G}_t,\mathbf{h}_t)$,
with two additional cross-summaries $\mathbf{G}_t,\mathbf{h}_t$ that enforce strict causality at second order.
(C) The chunk-wise parallel form interpolates between (A) and (B): the sequence is split into $M$ chunks of size $C$, intra-chunk computation evaluates the masked second-order form in parallel.}
\label{fig:hla_equivalence}
\end{figure}

In this section, we will introduce Higher-order Linear Attention (\textbf{HLA}). We begin with second-order linear attention as a warm-up, and present its extension to third-order linear attention in Section~\ref{sec:third-order-linear-attention}. 

\medskip\noindent\textbf{Second-order tensor attention mechanism.}
Second-order tensor attention can be written as
\[
\mathbf{T}_{2} := (\mathbf{Q}\mathbf{K}^\top)(\mathbf{Q}\mathbf{K}^\top)^\top=\mathbf{Q}(\mathbf{K}^\top\mathbf{K})\mathbf{Q}^\top \in 
\RR^{n\times n}, 
\]
\[
\operatorname{HLA}_2(\Qb, \Kb, \Vb) = \mathbf{T}_{2} \Vb \in \RR^{n \times d},
\]
so that $[\mathbf{T}_{2}]_{ij}=\mathbf{q}_i^\top(\mathbf{K}^\top\mathbf{K})\mathbf{q}_j$. The right-hand side shows a dependence on the second moment $\mathbf{K}^\top\mathbf{K}\in\mathbb{R}^{d\times d}$, suggesting streaming implementations via prefix moments.

We maintain prefix summaries at time $t$:
\begin{align*}
\mathbf{S}_t^K &\coloneqq \sum_{i\le t}\mathbf{k}_i\mathbf{k}_i^\top \in \RR^{d\times d}, \\
\mathbf{C}_t^{QV} &\coloneqq \sum_{i\le t}\mathbf{q}_i\mathbf{v}_i^\top \in \RR^{d\times d_v}, \\
\mathbf{m}_t^Q &\coloneqq \sum_{i\le t}\mathbf{q}_i \in \RR^{d}.
\end{align*}
All the above prefix summaries can be updated in a streaming fashion. In particular, the updates of $\mathbf{S}_t^K$ and $\mathbf{C}_t^{QV}$ cost $O(d^2)$ and $O(d\,d_v)$ time per token, respectively.

\medskip\noindent\textbf{Unnormalized HLA.}
The output of second-order HLA at time $t$ is the numerator-style bilinear form built from prefix moments:
\begin{equation}
\label{eq:hla-unnorm}
\mathbf{o}_t \;\coloneqq\; \mathbf{q}_t^\top \mathbf{S}_t^K \mathbf{C}_t^{QV} .
\end{equation}
This choice avoids length-dependent renormalization while preserving streaming updates and the same state as the normalized variant. 

\medskip\noindent\textbf{Normalized HLA.}
In order to define the normalized output of HLA, we define the numerator and denominator at $t$ as follows:
\[
\text{num}_t=\mathbf{q}_t^\top \mathbf{S}_t^K \mathbf{C}_t^{QV},\qquad
\text{den}_t=\mathbf{q}_t^\top \mathbf{S}_t^K \mathbf{m}_t^Q,
\]
and the normalized output of HLA is given by
\begin{equation}
\label{eq:hla-row}
\mathbf{o}_t= \frac{\text{num}_t}{\text{den}_t+\varepsilon}= \frac{\mathbf{q}_t^\top \mathbf{S}_t^K \mathbf{C}_t^{QV}}{\mathbf{q}_t^\top \mathbf{S}_t^K \mathbf{m}_t^Q+\varepsilon},
\end{equation}
where $\varepsilon >0$ is a small constant added for numerical stability.

Notably, $\mathbf{S}_t^K$ acts as a learned, data-dependent metric on query space; $\mathbf{C}_t^{QV}$ is a value accumulator modulated by past queries; and $\mathbf{m}_t^Q$ provides a query mass for optional scale control. This mirrors a second-order polynomial kernel in $(\mathbf{q},\mathbf{k})$ while remaining strictly streaming and causal once masked (Section~\ref{sec:mask}).

\medskip\noindent\textbf{Connection with linear attention.}
Setting $\mathbf{S}_t^K=\mathbf{I}$ yields
\[
\text{num}_t=\mathbf{q}_t^\top \mathbf{C}_t^{QV}
=\sum_{i\le t}(\mathbf{q}_t^\top \mathbf{q}_i)\,\mathbf{v}_i^\top,
\qquad
\text{den}_t=\mathbf{q}_t^\top \mathbf{m}_t^Q=\sum_{i\le t}\mathbf{q}_t^\top\mathbf{q}_i,
\]
So the normalized output reduces to a linear-attention form with kernel $K(\mathbf{q}_t,\mathbf{q}_i)=\mathbf{q}_t^\top \mathbf{q}_i$. When queries and keys are tied ($\mathbf{q}_i\equiv \mathbf{k}_i$), this coincides with linear attention using the identity feature map $\phi(x)=x$. In general, second-order HLA implements the \emph{data-adaptive} degree-2 polynomial kernel
$
K_t(\mathbf{q},\mathbf{q}')=\mathbf{q}^\top \mathbf{S}_t^K \mathbf{q}'
$
whose metric $\mathbf{S}_t^K=\sum_{i\le t}\mathbf{k}_i\mathbf{k}_i^\top$ depends on the past keys, strictly enriching first-order linearizations while retaining streaming. Absent tying $\mathbf{q}\equiv\mathbf{k}$, this differs from identity-feature linear attention.

\subsection{Causal masking via extended summaries}
\label{sec:mask}
Let $\mathbf{L}$ denote the binary causal mask (lower-triangular, including the diagonal). For the masked second-order matrix,
\[
\big[(\mathbf{L}\odot \mathbf{Q}\mathbf{K}^\top)(\mathbf{L}\odot \mathbf{Q}\mathbf{K}^\top)^\top\big]_{t,j}
=\sum_{i\le \min(t,j)}(\mathbf{q}_t^\top \mathbf{k}_i)(\mathbf{q}_j^\top \mathbf{k}_i)
=\mathbf{q}_t^\top \mathbf{S}_{\min(t,j)}^K \mathbf{q}_j.
\]

Equivalently, the strictly causal second-order output at time $t$ can be written in matrix form by masking on the right before applying values:
\[
\mathbf{o}_t
=\Big(\big[(\mathbf{L}\odot \mathbf{Q}\mathbf{K}^\top)(\mathbf{L}\odot \mathbf{Q}\mathbf{K}^\top)^\top\big]\odot \mathbf{L}\Big)_{t,:}\,\mathbf{V}.
\]
This row-wise $\odot \mathbf{L}$ enforces the restriction $j\le t$ when multiplying by $\mathbf{V}$.

Define two additional prefix summaries
\begin{align*}
\mathbf{G}_t &\coloneqq \sum_{i\le t} \left(\mathbf{k}_i\mathbf{k}_i^\top\right) \mathbf{C}_{i-1}^{QV}\in\RR^{d\times d_v},\\
\mathbf{h}_t &\coloneqq \sum_{i\le t} \left(\mathbf{k}_i\mathbf{k}_i^\top\right) \mathbf{m}_{i-1}^{Q}\in\RR^{d}.
\end{align*}
We have the following theorem, which gives the unnormalized and normalized outputs of HLA with a causal mask.

\begin{theorem}[Masked streaming identity for second order]
\label{thm:masked-second}
For each $t$, let
\[
\text{num}_t^{\mathrm{mask}}=\mathbf{q}_t^\top\!\left(\mathbf{S}_t^K \mathbf{C}_t^{QV}-\mathbf{G}_t\right),\qquad
\text{den}_t^{\mathrm{mask}}=\mathbf{q}_t^\top\!\left(\mathbf{S}_t^K \mathbf{m}_t^{Q}-\mathbf{h}_t\right).
\]
Consequently, the strictly causal, masked default unnormalized output is
\begin{equation}
\label{eq:masked-unnorm}
\mathbf{o}_t \;=\; \mathbf{q}_t^\top\!\left(\mathbf{S}_t^K \mathbf{C}_t^{QV}-\mathbf{G}_t\right).
\end{equation}
An optional linear normalization divides by the masked denominator,
\begin{equation}
\label{eq:masked}
\mathbf{o}_t=
\frac{\mathbf{q}_t^\top\!\left(\mathbf{S}_t^K \mathbf{C}_t^{QV}-\mathbf{G}_t\right)}
{\mathbf{q}_t^\top\!\left(\mathbf{S}_t^K \mathbf{m}_t^{Q}-\mathbf{h}_t\right)+\varepsilon},
\end{equation}
where $\varepsilon >0$ is a small constant added for numerical stability.
\end{theorem}

\begin{proof}
Let $\mathbf{W}=\mathbf{L}\odot(\mathbf{Q}\mathbf{K}^\top)$ with $\mathbf{L}$ lower-triangular including the diagonal. For the second-order weight matrix, we have
$\mathbf{W}\mathbf{W}^\top$ with entries
$(\mathbf{W}\mathbf{W}^\top)_{t,j}=\sum_{i\le \min(t,j)}(\mathbf{q}_t^\top\mathbf{k}_i)(\mathbf{q}_j^\top\mathbf{k}_i)$.
Then the masked, unnormalized numerator at time $t$ is
\[
\text{num}_t^{\mathrm{mask}}=\sum_{j\le t}(\mathbf{W}\mathbf{W}^\top)_{t,j}\,\mathbf{v}_j^\top
\,=\,\sum_{j\le t}\Big(\sum_{i\le j} \mathbf{q}_t^\top\mathbf{k}_i\,\mathbf{k}_i^\top\mathbf{q}_j\Big)\mathbf{v}_j^\top
=\mathbf{q}_t^\top\sum_{j\le t}\!\Big(\sum_{i\le j}\mathbf{k}_i\mathbf{k}_i^\top\Big)\mathbf{q}_j\mathbf{v}_j^\top,
\]
where the second equality uses the fact that $\min(t,j) = j$ when $j\le t$.
Interchanging finite sums yields
\begin{align}
\sum_{j\le t}\!\Big(\sum_{i\le j}\mathbf{k}_i\mathbf{k}_i^\top\Big)\mathbf{q}_j\mathbf{v}_j^\top
=\sum_{j\le t}\mathbf{S}_j^K\,\mathbf{q}_j\mathbf{v}_j^\top
=\underbrace{\sum_{j\le t}\mathbf{S}_t^K\,\mathbf{q}_j\mathbf{v}_j^\top}_{I_1}
- \underbrace{\sum_{j\le t}\Big(\sum_{j<i\le t} \mathbf{k}_i\mathbf{k}_i^\top\Big)\mathbf{q}_j\mathbf{v}_j^\top}_{I_2},\label{eq:Gu0001}
\end{align}
where the last equality holds due to $\mathbf{S}_j^K = \mathbf{S}_t^K - \sum_{j<i\le t} \mathbf{k}_i\mathbf{k}_i^\top$.

In Eq.~\eqref{eq:Gu0001}, the first term $I_1$ equals $\mathbf{S}_t^K \mathbf{C}_t^{QV}$. For the second term $I_2$, swap the order of summation:
$\sum_{j\le t}\sum_{i>j}(\cdot)=\sum_{i\le t}\sum_{j<i}(\cdot)$, we can obtain
$I_2= \sum_{i\le t}\left(\mathbf{k}_i\mathbf{k}_i^\top\right)\!\!\left(\sum_{j<i}\mathbf{q}_j\mathbf{v}_j^\top\right)=\mathbf{G}_t$. 
This proves the numerator identity. The proof for the denominator is analogous with $\mathbf{v}_j$ replaced by $1$ (i.e., $\mathbf{q}_j$ replaced by $1$-summaries), yielding
$\mathbf{S}_t^K \mathbf{m}_t^Q - \mathbf{h}_t$. Finally, the division by $\text{den}_t^{\mathrm{mask}}+\varepsilon$ gives Eq.~\eqref{eq:masked}.
\end{proof}

\medskip\noindent\textbf{Online updates.}
Using the fact that $(\mathbf{k}\mathbf{k}^\top)X=\mathbf{k}(\mathbf{k}^\top X)$, we have
\[
\begin{aligned}
\mathbf{S}_t^K&=\mathbf{S}_{t-1}^K+\mathbf{k}_t\mathbf{k}_t^\top, &
\mathbf{C}_t^{QV}&=\mathbf{C}_{t-1}^{QV}+\mathbf{q}_t\mathbf{v}_t^\top, &
\mathbf{m}_t^Q&=\mathbf{m}_{t-1}^Q+\mathbf{q}_t,\\
\mathbf{G}_t&=\mathbf{G}_{t-1}+\mathbf{k}_t(\mathbf{k}_t^\top \mathbf{C}_{t-1}^{QV}), &
\mathbf{h}_t&=\mathbf{h}_{t-1}+\mathbf{k}_t(\mathbf{k}_t^\top \mathbf{m}_{t-1}^{Q}). &&
\end{aligned}
\]
Therefore, the per-token cost remains $O(d^2{+}d\,d_v)$ in total.

\section{Chunk-parallel training via associative scans}
\label{sec:scan}
In Section \ref{sec:mask}, we have presented the recurrent form for second-order HLA. As we know, training a purely recurrent model is inefficient on GPUs. We adopt within-chunk scans with width $w$ and inter-chunk scans across $B$ chunks~\citep{blelloch1990prefix}. A similar technique has been widely used in the literature of linear attention~\citep{yang2023gated,qin2024lightning}.
We write $B_c$ for the number of chunks to avoid overloading $B$ elsewhere; thus, inter-chunk scans are across $B_c$ chunks.

\subsection{Unmasked monoid}
Let $\mathcal{S}=(\mathbf{S},\mathbf{C},\mathbf{m})$ with token “deltas’’ $\Delta \mathbf{S}_t=\mathbf{k}_t\mathbf{k}_t^\top$, $\Delta \mathbf{C}_t=\mathbf{q}_t\mathbf{v}_t^\top$, $\Delta \mathbf{m}_t=\mathbf{q}_t$. Define elementary segments $\mathcal{T}_t=(\Delta \mathbf{S}_t,\Delta \mathbf{C}_t,\Delta \mathbf{m}_t)$ and the additive monoid
\[
(\mathbf{S}_A,\mathbf{C}_A,\mathbf{m}_A)\oplus(\mathbf{S}_B,\mathbf{C}_B,\mathbf{m}_B)=(\mathbf{S}_A{+}\mathbf{S}_B,\, \mathbf{C}_A{+}\mathbf{C}_B,\, \mathbf{m}_A{+}\mathbf{m}_B).
\]
An exclusive Blelloch scan on $\{\mathcal{T}_1,\ldots,\mathcal{T}_w\}$ yields per-token prefixes $\mathcal{P}_t=\bigoplus_{i<t}\mathcal{T}_i$, from which the inclusive state at $t$ is obtained locally by adding $\mathcal{T}_t$. Here and below, $A$ then $B$ denotes adjacent segments in time (all indices in $A$ precede those in $B$).

\subsection{Masked semidirect product}
For the masked case use $\mathcal{S}=(\mathbf{S},\mathbf{C},\mathbf{m},\mathbf{G},\mathbf{h})$. For a single-token segment, $\mathbf{G}=\mathbf{h}=\mathbf{0}$. Concatenation is
\begin{equation}
\label{eq:semidirect}
\begin{aligned}
(\mathbf{S}_A,&\mathbf{C}_A,\mathbf{m}_A,\mathbf{G}_A,\mathbf{h}_A)\oplus(\mathbf{S}_B,\mathbf{C}_B,\mathbf{m}_B,\mathbf{G}_B,\mathbf{h}_B)=\\
&\big(\mathbf{S}_A{+}\mathbf{S}_B,\; \mathbf{C}_A{+}\mathbf{C}_B,\; \mathbf{m}_A{+}\mathbf{m}_B,\; \mathbf{G}_A{+}\mathbf{G}_B+\mathbf{S}_B \mathbf{C}_A,\; \mathbf{h}_A{+}\mathbf{h}_B+\mathbf{S}_B \mathbf{m}_A\big),
\end{aligned}
\end{equation}
which is associative (direct expansion). Perform the same exclusive scan; per-token inclusive states follow by adding the local deltas and the cross-terms $\Delta \mathbf{S}_t \mathbf{C}_{t-1}$ and $\Delta \mathbf{S}_t \mathbf{m}_{t-1}$.

\vspace{0.5em}
\noindent\textbf{Decay-aware monoid.}
Let $\gamma\in(0,1)$ be a fixed exponential decay and let a segment $\mathcal{X}$ carry its \emph{length} $\ell(\mathcal{X})$ and attenuation $\rho(\mathcal{X})\coloneqq \gamma^{\ell(\mathcal{X})}$. For the unmasked triple $\mathcal{S}=(\mathbf{S},\mathbf{C},\mathbf{m})$ the decayed concatenation is
\[
(\mathbf{S}_A,\mathbf{C}_A,\mathbf{m}_A,\rho_A)\oplus_\gamma(\mathbf{S}_B,\mathbf{C}_B,\mathbf{m}_B,\rho_B)
=\big(\rho_B \mathbf{S}_A{+}\mathbf{S}_B,\;\rho_B \mathbf{C}_A{+}\mathbf{C}_B,\;\rho_B \mathbf{m}_A{+}\mathbf{m}_B,\;\rho_A\rho_B\big),
\]
and analogously for the masked $(\mathbf{S},\mathbf{C},\mathbf{m},\mathbf{G},\mathbf{h})$ state:
\[
\begin{aligned}
(\mathbf{S}_A,&\mathbf{C}_A,\mathbf{m}_A,\mathbf{G}_A,\mathbf{h}_A,\rho_A)\oplus_\gamma(\mathbf{S}_B,\mathbf{C}_B,\mathbf{m}_B,\mathbf{G}_B,\mathbf{h}_B,\rho_B)=\\
&\Big(\rho_B \mathbf{S}_A{+}\mathbf{S}_B,\;\rho_B \mathbf{C}_A{+}\mathbf{C}_B,\;\rho_B \mathbf{m}_A{+}\mathbf{m}_B,\\
&\qquad \rho_B \mathbf{G}_A{+}\mathbf{G}_B+\mathbf{S}_B(\rho_B \mathbf{C}_A),\;\rho_B \mathbf{h}_A{+}\mathbf{h}_B+\mathbf{S}_B(\rho_B \mathbf{m}_A),\;\rho_A\rho_B\Big).
\end{aligned}
\]
Associativity follows from bilinearity and $\rho$-multiplicativity.

\begin{theorem}[Scan equivalence: serial vs.\ (decayed) associative scans]
\label{thm:scan-equivalence}
Consider a sequence of token segments $\{\mathcal{T}_1,\ldots,\mathcal{T}_n\}$ and either $\oplus$ (no decay) or $\oplus_\gamma$ (with decay). Let $\mathcal{P}_t$ be the exclusive prefix obtained by a Blelloch scan under the chosen operator. For each $t$, the inclusive state computed locally from $\mathcal{P}_t$ and $\mathcal{T}_t$ equals the state produced by a serial left-to-right recurrence on tokens $1{:}t$. Consequently, the per-token masked outputs are identical to those of the serial algorithm.
\end{theorem}
\begin{proof}
We prove for the masked, decayed case; the other cases are specializations. Define the serial recurrence
$\mathcal{X}_t=\Phi_\gamma(\mathcal{X}_{t-1},\mathcal{T}_t)$ given by the online updates in Section~\ref{sec:mask} with decay $\gamma$. By construction,
$\Phi_\gamma$ coincides with the binary map $f_\gamma(\mathcal{X},\mathcal{Y})\coloneqq \mathcal{X}\oplus_\gamma \mathcal{Y}$ when $\mathcal{Y}$ is a single-token segment. Because $\oplus_\gamma$ is associative with identity the zero-length segment $\mathcal{E}$ (all-zero summaries, $\rho=1$), the Blelloch scan yields
$\mathcal{P}_t=\mathcal{E}\oplus_\gamma \mathcal{T}_1 \oplus_\gamma \cdots \oplus_\gamma \mathcal{T}_{t-1}$. The local inclusive update computes $\mathcal{P}_t\oplus_\gamma \mathcal{T}_t$, which equals $\mathcal{X}_t$ by associativity and the definition of $\Phi_\gamma$. The masked outputs are functions only of the inclusive state (Theorem~\ref{thm:masked-second}), hence coincide with the serial outputs.
\end{proof}

\medskip\noindent\textbf{Backward for gradients.}
Let $\oplus^\ast$ denote the vector-Jacobian adjoint of $\oplus$ evaluated at the forward states. A reverse (decayed) scan applying $\oplus_\gamma^\ast$ with checkpointing at tile boundaries yields gradients that match those of the serial recurrence, by Theorem~\ref{thm:scan-equivalence} and the chain rule. 

\begin{remark}[Inclusive vs.\ exclusive scans]
\label{rem:scan-modes}

Given segments $(\mathcal{T}_1,\ldots,\mathcal{T}_w)$ and an associative operator $\oplus$ with identity $\mathcal{E}$, the \emph{exclusive} scan returns prefixes
$\mathcal{P}_t=\mathcal{E}\oplus\mathcal{T}_1\oplus\cdots\oplus\mathcal{T}_{t-1}$,
while the \emph{inclusive} scan returns
$\mathcal{I}_t=\mathcal{P}_t\oplus\mathcal{T}_t$.
Our forward algorithms compute $\mathcal{P}_t$ via an exclusive Blelloch scan and then form the inclusive state locally by combining $\mathcal{P}_t$ with the token’s deltas (and required cross-terms). This choice exposes maximal parallelism and ensures exact equality to the serial recurrence by Theorem~\ref{thm:scan-equivalence}. With decay, the identity is the zero-length segment $(\mathbf{0},\ldots,\mathbf{0},\rho{=}1)$; the exclusive/inclusive distinction is unchanged.

\textbf{Intra-chunk parallelism.} Within a chunk of width $w$, an exclusive Blelloch scan over $\{\mathcal{T}_1,\ldots,\mathcal{T}_w\}$ under $\oplus$ (or $\oplus_\gamma$) yields $\mathcal{P}_t$ for all $t$ in $O(\log w)$ span and $O(1)$ auxiliary memory per position. The per-token inclusive states are then computed independently as $\mathcal{I}_t=\mathcal{P}_t\oplus\mathcal{T}_t$.

\textbf{Inter-chunk parallelism.} For $B_c$ chunks, each chunk $c$ produces a single \emph{summary} $\mathcal{S}^{(c)}=\bigoplus_{t\in\text{chunk }c}\mathcal{T}_t$. An exclusive scan across the $B_c$ summaries gives carry-in prefixes $\widehat{\mathcal{P}}^{(c)}$ for every chunk. Each position $t$ in chunk $c$ then uses the merged prefix $\widehat{\mathcal{P}}^{(c)}\oplus \mathcal{P}_t$ before adding its local $\mathcal{T}_t$ to obtain the inclusive state. This is the same parallel skeleton widely used in modern linear-attention and recurrent networks that maintain streaming sufficient statistics~\citep{sun2023retentive, qin2023transnormerllm, yang2023gated, qin2024lightning, yang2024parallelizing}.

\textbf{Connection to linear attention.} First-order linear attentions and related modern RNN kernels scan additive/decayed summaries (e.g., $\sum\phi(\mathbf{k})\mathbf{v}^\top$ and denominators) using exactly this intra-/inter-chunk pattern. \textbf{HLA} plugs into the same infrastructure: only the state tuple and cross-terms change (e.g., $(\mathbf{S},\mathbf{C},\mathbf{m},\mathbf{G},\mathbf{h})$ for second order), while the exclusive/inclusive logic and two-level scan strategy remain identical. Thus, HLA inherits the throughput characteristics of these systems with strictly higher expressivity.
\end{remark}

\subsection{Adding decay and regularization}
\label{sec:decay}
\medskip\noindent\textbf{Decayed states.}
Introduce a time decay $\gamma\in(0,1)$:
\[
\mathbf{S}_t^K=\gamma \mathbf{S}_{t-1}^K+\mathbf{k}_t\mathbf{k}_t^\top,\quad
\mathbf{C}_t^{QV}=\gamma \mathbf{C}_{t-1}^{QV}+\mathbf{q}_t\mathbf{v}_t^\top,\quad
\mathbf{m}_t^{Q}=\gamma \mathbf{m}_{t-1}^{Q}+\mathbf{q}_t,
\]
and the cross-summaries obey
\[
\mathbf{G}_t=\gamma \mathbf{G}_{t-1}+\mathbf{k}_t\big(\mathbf{k}_t^\top \mathbf{C}_{t-1}^{QV}\big),\qquad
\mathbf{h}_t=\gamma \mathbf{h}_{t-1}+\mathbf{k}_t\big(\mathbf{k}_t^\top \mathbf{m}_{t-1}^{Q}\big),
\]
which are the decayed analogues of the online updates in Section~\ref{sec:mask}. Decay controls spectral growth and improves recency bias while maintaining associativity (with respect to segment-local normalization)~\citep{peng2023rwkv, sun2023retentive, qin2023transnormerllm, yang2023gated, yang2024gated, peng2024eagle, behrouz2024titans, peng2025rwkv, behrouz2025atlas, behrouz2025s}.

\section{Implementation details and complexity}
In this section, we discuss the implementation details and provide a complexity analysis.

Recall that for each token and each head (second order), we have
\begin{itemize}
\item \textbf{State:} $\mathbf{S}_t^K\in\RR^{d\times d}$, $\mathbf{C}_t^{QV}\in\RR^{d\times d_v}$, $\mathbf{m}_t^Q\in\RR^d$ (and masked $\mathbf{G}_t\in\RR^{d\times d_v}$, $\mathbf{h}_t\in\RR^d$).
\item \textbf{Compute:} evaluate $\mathbf{u}_t=\mathbf{q}_t^\top \mathbf{S}_t^K$ (mat–vec) and then $\mathbf{u}_t \mathbf{C}_t^{QV}$ (row–matrix), with masked corrections $-\mathbf{q}_t^\top\mathbf{G}_t$; the denominator uses $\mathbf{u}_t\mathbf{m}_t^Q-\mathbf{q}_t^\top \mathbf{h}_t$. This avoids forming $\mathbf{S}_t^K\mathbf{C}_t^{QV}$ explicitly; masked cross-terms still use $\mathbf{k}_t^\top X$ to avoid cubic cost.
\item \textbf{Parallelism:} within-chunk Blelloch scans (span $O(\log w)$) and inter-chunk exclusive scans across $B_c$ chunks, both using the same $\oplus$.
\end{itemize}

\begin{algorithm}[H]
\small
\caption{Masked (Second Order) HLA with Within-Chunk Scan}
\label{alg:masked-hla}
\begin{algorithmic}[1]
\Require Chunk tokens $(\mathbf{q}[1{:}w],\mathbf{k}[1{:}w],\mathbf{v}[1{:}w])$, $\varepsilon$, optional ridge $\lambda$, optional decay $\gamma$, optional flag \texttt{normalize}
\State \textbf{Token segments:} for $t=1..w$, set $\Delta \mathbf{S}_t \gets \mathbf{k}_t \mathbf{k}_t^\top$, $\Delta \mathbf{C}_t \gets \mathbf{q}_t \mathbf{v}_t^\top$, $\Delta \mathbf{m}_t \gets \mathbf{q}_t$, and initialize $\mathbf{G}_t{=}\mathbf{0}$, $\mathbf{h}_t{=}\mathbf{0}$.
\State \textbf{Exclusive scan} over $\{(\Delta \mathbf{S}_t,\Delta \mathbf{C}_t,\Delta \mathbf{m}_t,\mathbf{0},\mathbf{0})\}_{t=1}^w$ using $\oplus$ in Eq.~\eqref{eq:semidirect} (with decay if used) to obtain prefixes $\mathcal{P}_t=(\mathbf{S}_{t-1},\mathbf{C}_{t-1},\mathbf{m}_{t-1},\mathbf{G}_{t-1},\mathbf{h}_{t-1})$.
\For{$t=1$ to $w$ \textbf{in parallel}}
    \State \textbf{Inclusive state:}
    \State $\mathbf{S}_t \gets \gamma \mathbf{S}_{t-1} + \Delta \mathbf{S}_t$;\quad $\mathbf{C}_t \gets \gamma \mathbf{C}_{t-1} + \Delta \mathbf{C}_t$;\quad $\mathbf{m}_t \gets \gamma \mathbf{m}_{t-1} + \Delta \mathbf{m}_t$
    \State $\mathbf{G}_t \gets \gamma \mathbf{G}_{t-1} + \Delta \mathbf{S}_t\, \mathbf{C}_{t-1}$;\quad $\mathbf{h}_t \gets \gamma \mathbf{h}_{t-1} + \Delta \mathbf{S}_t\, \mathbf{m}_{t-1}$
    \State \textbf{Effective $\mathbf{S}$:} $\mathbf{S}_t^{\mathrm{eff}} \gets \mathbf{S}_t + \lambda \mathbf{I}$ \Comment{optional ridge for stability}
    \State \textbf{Default masked unnormalized output:}
    \State $\mathbf{u} \gets \mathbf{q}_t^\top \mathbf{S}_t^{\mathrm{eff}}$ \Comment{$O(d^2)$ matvec}
    \State $\text{num}\gets \mathbf{u}\,\mathbf{C}_t - \mathbf{q}_t^\top \mathbf{G}_t$ \Comment{$O(d\,d_v)$}
    \State $\mathbf{o}_t^{\mathrm{hla}} \gets \text{num}$
    \State \textbf{Optional normalization:}
    \If{\texttt{normalize}}
        \State $\text{den}\gets \mathbf{u}\,\mathbf{m}_t - \mathbf{q}_t^\top \mathbf{h}_t + \varepsilon$
        \State $\mathbf{o}_t^{\mathrm{hla}} \gets \mathbf{o}_t^{\mathrm{hla}} / \text{den}$
    \EndIf
\EndFor
\State \Return $\{\mathbf{o}_t^{\mathrm{hla}}\}_{t=1}^w$
\end{algorithmic}
\end{algorithm}

\subsection{Pseudocode}
We present a PyTorch-like reference for masked second-order HLA with a within-chunk exclusive scan. Unmasked and/or diagonal-regularized variants follow by removing $(\mathbf{G},\mathbf{h})$. Normalization is optional; by default, the implementation returns the unnormalized output and may divide by the masked denominator if requested. 

\noindent\textbf{Remark.} Adding $\lambda\mathbf{I}$ yields a stabilized causal variant of the masked operator; it does not correspond to the exact masked bilinear form of $(\mathbf{L}\odot \mathbf{Q}\mathbf{K}^\top)$.

\subsection{Implementation considerations}
\label{sec:impl}

HLA only replaces the standard attention sublayer in the transformer block, while the feed-forward sublayer and normalization sublayers remain unchanged. Drop-in replacement requires only swapping the kernel while keeping positional encodings and masking identical to the baseline. Multi-query keys/values (sharing $\mathbf{K},\mathbf{V}$ across heads) reduce state from $O(h\,d^2)$ to $O(d^2{+}h\,d\,d_v)$ while preserving the algebra.

The summaries $(\mathbf{S},\mathbf{C},\mathbf{m},\mathbf{G},\mathbf{h})$ are per head. With \emph{multi-query} ($\mathbf{K},\mathbf{V}$ shared across heads), the key moment $\mathbf{S}_t^K$ is shared and stored once per layer ($O(d^2)$), while $(\mathbf{C}_t^{QV},\mathbf{m}_t^Q,\mathbf{G}_t,\mathbf{h}_t)$ remain per-head ($O(h\,d\,d_v{+}h\,d)$). This yields a total memory of $O(d^2{+}h\,d\,d_v)$ instead of $O(h\,d^2{+}h\,d\,d_v)$ when each head maintains its own $\mathbf{S}_t^K$.

For throughput, maintain $\mathbf{S}_t^K$ in a packed symmetric layout (store only the upper triangle, $\tfrac{1}{2}d(d{+}1)$ entries) to reduce bandwidth without changing the algebra. Within a chunk of width $w$, use an exclusive Blelloch scan to obtain prefixes in $O(\log w)$ span and constant extra memory per position; inter-chunk scans use the same operator across $B_c$ chunks.

\section{Asymmetric Higher-order Linear Attention}
\label{sec:AHLA}

%
\begin{figure}[ht!]
\centering
\resizebox{0.8\linewidth}{!}{%
\begin{tikzpicture}[
font=\sffamily,
>=Latex,
base_node/.style={
  thick,
  rounded corners=8pt,
  blur shadow={shadow blur steps=5, shadow opacity=15}
},
tensor/.style={
  base_node,
  draw=appleBlue,
  fill=appleLightBlue,
  minimum height=2.4em,
  minimum width=2.4em,
  font=\sffamily\bfseries
},
value_tensor/.style={
  base_node,
  draw=appleGreen,
  fill=appleLightGreen,
  minimum height=2.4em,
  minimum width=2.4em,
  font=\sffamily\bfseries
},
query_tensor/.style={
  base_node,
  draw=appleRed,
  fill=appleLightRed,
  minimum height=2.4em,
  minimum width=2.4em,
  font=\sffamily\bfseries
},
state/.style={
  base_node,
  draw=appleGray,
  fill=appleLightGray,
  rounded corners=10pt,
  minimum height=3.6em,
  minimum width=3.6em,
  align=center,
  font=\sffamily\bfseries
},
cross_state/.style={
  base_node,
  draw=appleTeal,
  fill=appleLightTeal,
  rounded corners=10pt,
  minimum height=3.6em,
  minimum width=3.6em,
  align=center,
  font=\sffamily\bfseries
},
op/.style={
  circle,
  fill=appleLightGray,
  draw=appleGray,
  inner sep=2pt,
  thick,
  font=\sffamily\bfseries\small
},
arrow/.style={
  ->,
  thick,
  color=appleDarkGray,
  line width=1.2pt,
  rounded corners=4pt
},
update_arrow/.style={
  arrow,
  color=appleGreen!65!black
},
cross_arrow/.style={
  arrow,
  color=appleTeal!75!black,
  dashed,
  line width=1.0pt
},
read_arrow/.style={
  arrow,
  color=appleRed
},
section_label/.style={
  font=\sffamily\bfseries\Large,
  anchor=north west,
  color=appleDark
},
equiv_badge/.style={
  base_node,
  draw=appleDark,
  fill=white,
  rounded corners=14pt,
  inner sep=8pt,
  minimum width=1.6cm,
  minimum height=1.6cm,
  font=\sffamily\bfseries\Huge,
  text=appleDark
},
eqbox/.style={
  base_node,
  fill=white,
  draw=appleGray!50,
  inner sep=9pt
}
]


\begin{scope}[local bounding box=panelA]

\node[font=\sffamily\bfseries\small, text=appleGray] (timeLabelA) at (-1.6,3.0) {time $\rightarrow$};

\node[state] (S0)
  {$\bigl(\mathbf{P}^{KV}_{0},\mathbf{m}^{K}_{0},$\\$\mathbf{E}_{0},\mathbf{n}_{0}\bigr)$};
\node[state, right=2.0cm of S0] (S1)
  {$\bigl(\mathbf{P}^{KV}_{1},\mathbf{m}^{K}_{1},$\\$\mathbf{E}_{1},\mathbf{n}_{1}\bigr)$};
\node[state, right=2.0cm of S1] (S2)
  {$\bigl(\mathbf{P}^{KV}_{2},\mathbf{m}^{K}_{2},$\\$\mathbf{E}_{2},\mathbf{n}_{2}\bigr)$};
\node[font=\sffamily\bfseries, right=0.8cm of S2] (sdots) {$\cdots$};
\node[state, right=0.8cm of sdots] (St)
  {$\bigl(\mathbf{P}^{KV}_{t},\mathbf{m}^{K}_{t},$\\$\mathbf{E}_{t},\mathbf{n}_{t}\bigr)$};

\node[query_tensor, scale=0.8, above=1.1cm of S1, xshift=-14pt] (q1) {$\mathbf{q}_1$};
\node[tensor, scale=0.8, right=0.05cm of q1] (k1) {$\mathbf{k}_1$};
\node[value_tensor, scale=0.8, right=0.05cm of k1] (v1) {$\mathbf{v}_1$};

\node[query_tensor, scale=0.8, above=1.1cm of S2, xshift=-14pt] (q2) {$\mathbf{q}_2$};
\node[tensor, scale=0.8, right=0.05cm of q2] (k2) {$\mathbf{k}_2$};
\node[value_tensor, scale=0.8, right=0.05cm of k2] (v2) {$\mathbf{v}_2$};

\node[query_tensor, scale=0.8, above=1.1cm of St, xshift=-14pt] (qt) {$\mathbf{q}_t$};
\node[tensor, scale=0.8, right=0.05cm of qt] (kt) {$\mathbf{k}_t$};
\node[value_tensor, scale=0.8, right=0.05cm of kt] (vt) {$\mathbf{v}_t$};

\draw[arrow] (S0) -- (S1);
\draw[arrow] (S1) -- (S2);
\draw[arrow] (S2) -- (sdots);
\draw[arrow] (sdots) -- (St);

\draw[update_arrow] ($(q1.south)!0.5!(v1.south)$) -- (S1.north);
\draw[update_arrow] ($(q2.south)!0.5!(v2.south)$) -- (S2.north);
\draw[update_arrow] ($(qt.south)!0.5!(vt.south)$) -- (St.north);

\node[font=\sffamily\bfseries\scriptsize, text=appleGreen!50!black, align=center]
  at ($(S0)+(0.0,-1.20)$) {write outer products\\$+$ routed term $\mathbf{k}_t(\mathbf{q}_t^{\!\top}\mathbf{P}^{KV}_{t})$};

\node[query_tensor, scale=0.9, below=1.2cm of St] (qtR) {$\mathbf{q}_t$};
\node[base_node, dashed, draw=appleGray, fill=white,
      minimum width=1.7cm, minimum height=2.2em,
      right=1.3cm of St] (ytA) {$\mathbf{o}_t$};

\draw[read_arrow] (qtR.north) -- (St.south);
\draw[arrow, color=appleRed!85!black] (St.east) -- (ytA.west);
\node[font=\sffamily\bfseries\scriptsize, text=appleRed,
      above=0.04cm of ytA, anchor=south]
  {read $\mathbf{q}_t^{\!\top}\mathbf{E}_t$};

\node[eqbox, below=2.4cm of S2, align=left, xshift=-0.6cm] (eqA) {%
$\displaystyle
\begin{aligned}
\mathbf{P}^{KV}_t &= \mathbf{P}^{KV}_{t-1} + \mathbf{k}_t\mathbf{v}_t^{\!\top},
\qquad
\mathbf{m}^{K}_t = \mathbf{m}^{K}_{t-1} + \mathbf{k}_t \\[2pt]
\mathbf{E}_t &= \mathbf{E}_{t-1} + \mathbf{k}_t\bigl(\mathbf{q}_t^{\!\top}\mathbf{P}^{KV}_{t}\bigr),
\quad
\mathbf{n}_t = \mathbf{n}_{t-1} + \mathbf{k}_t\bigl(\mathbf{q}_t^{\!\top}\mathbf{m}^{K}_{t}\bigr) \\[2pt]
\mathbf{o}_t &= \mathbf{q}_t^{\!\top}\,\mathbf{E}_t
\qquad\text{(optional norm: divide by }\mathbf{q}_t^{\!\top}\mathbf{n}_t + \varepsilon\text{)}
\end{aligned}$%
};

\node[base_node, fill=appleLightBlue!60, draw=appleBlue!50,
      inner sep=6pt, font=\sffamily\bfseries\scriptsize, text=appleBlue,
      right=0.55cm of eqA, align=center] (complA)
  {$\mathcal{O}(N\,d\,d_{v})$ time\\$\mathcal{O}(d\,d_{v})$ state at inference};

\end{scope}


\coordinate (xRefEquiv) at ($(panelA.south)+(-3.5,0)$);

\coordinate (yPosAB) at ($(panelA.south)+(0,-2.1)$);
\node[equiv_badge] (equiv) at (xRefEquiv |- yPosAB) {$\equiv$};

\node[eqbox, draw=appleDark!40, fill=appleLightGray!50,
      right=0.45cm of equiv, inner sep=8pt,
      font=\sffamily\small, text=appleDark, align=center] (derivBox)
{\textbf{exact equality for zero-initialised state}:\;
$\displaystyle
\mathbf{P}^{KV}_t=\!\!\sum_{j\le t}\!\mathbf{k}_j\mathbf{v}_j^{\!\top},\;\;
\mathbf{E}_t=\!\!\sum_{i\le t}\!\mathbf{k}_i\bigl(\mathbf{q}_i^{\!\top}\mathbf{P}^{KV}_{i}\bigr)$\\[2pt]
$\displaystyle
\Longrightarrow\;
\mathbf{o}_t \,=\, \mathbf{q}_t^{\!\top}\mathbf{E}_t
\,=\, \sum_{j\le t}\sum_{i=j}^{t}(\mathbf{q}_t^{\!\top}\mathbf{k}_i)(\mathbf{q}_i^{\!\top}\mathbf{k}_j)\,\mathbf{v}_j^{\!\top}
\,=\, \bigl[(\mathbf{A}\mathbf{A})\odot\mathbf{L}\bigr]_{t,:}\mathbf{V}$%
};


\begin{scope}[shift={(0,-12.2)}, local bounding box=panelB]

\node[base_node, draw=appleRed, fill=appleLightRed,
      minimum width=1.0cm, minimum height=2.4cm,
      align=center, font=\sffamily\bfseries] (Qmat) at (0,0)
  {$\mathbf{Q}$};
\node[font=\sffamily\scriptsize, text=appleRed, above=0.05cm of Qmat]
  {$N\!\times\!d$};

\node[op, right=0.45cm of Qmat] (mul1) {$\times$};

\node[base_node, draw=appleBlue, fill=appleLightBlue,
      minimum width=2.0cm, minimum height=1.0cm,
      align=center, font=\sffamily\bfseries, right=0.45cm of mul1] (KTmat)
  {$\mathbf{K}^{\!\top}$};
\node[font=\sffamily\scriptsize, text=appleBlue, above=0.05cm of KTmat]
  {$d\!\times\!N$};

\node[font=\sffamily\bfseries\Large, text=appleDark, right=0.45cm of KTmat] (eq1) {$=$};

\node[base_node, draw=appleGray, fill=appleLightGray,
      minimum width=2.0cm, minimum height=2.0cm,
      align=center, font=\sffamily\bfseries, right=0.5cm of eq1] (Amat) {};
\begin{scope}
  \clip[rounded corners=8pt] ([xshift=2pt,yshift=-2pt]Amat.north west)
        rectangle ([xshift=-2pt,yshift=2pt]Amat.south east);
  \fill[appleBlue!18]
    ([xshift=2pt,yshift=-2pt]Amat.north west) --
    ([xshift=-2pt,yshift=2pt]Amat.south east) --
    ([xshift=2pt,yshift=2pt]Amat.south west) -- cycle;
\end{scope}
\foreach \i in {0.2,0.5,0.8,1.1,1.4,1.7} {
  \draw[appleRed!22, line width=0.6pt]
    ([xshift=\i cm,yshift=-2pt]Amat.north west) --
    ([xshift=-2pt,yshift=-\i cm]Amat.north east);
}
\node[font=\sffamily\bfseries\scriptsize, text=appleDark, align=center]
  at (Amat.center) {$\mathbf{L}\odot\mathbf{Q}\mathbf{K}^{\!\top}$\\$\eqqcolon\mathbf{A}$};
\node[font=\sffamily\scriptsize, text=appleGray, above=0.05cm of Amat]
  {$N\!\times\!N$};

\node[op, right=0.45cm of Amat] (mul2) {$\times$};

\node[base_node, draw=appleGray, fill=appleLightGray,
      minimum width=2.0cm, minimum height=2.0cm,
      align=center, font=\sffamily\bfseries, right=0.45cm of mul2] (Amat2) {};
\begin{scope}
  \clip[rounded corners=8pt] ([xshift=2pt,yshift=-2pt]Amat2.north west)
        rectangle ([xshift=-2pt,yshift=2pt]Amat2.south east);
  \fill[appleBlue!18]
    ([xshift=2pt,yshift=-2pt]Amat2.north west) --
    ([xshift=-2pt,yshift=2pt]Amat2.south east) --
    ([xshift=2pt,yshift=2pt]Amat2.south west) -- cycle;
\end{scope}
\foreach \i in {0.2,0.5,0.8,1.1,1.4,1.7} {
  \draw[appleRed!22, line width=0.6pt]
    ([xshift=\i cm,yshift=-2pt]Amat2.north west) --
    ([xshift=-2pt,yshift=-\i cm]Amat2.north east);
}
\node[font=\sffamily\bfseries\scriptsize, text=appleDark, align=center]
  at (Amat2.center) {$\mathbf{A}$};
\node[font=\sffamily\scriptsize, text=appleGray, above=0.05cm of Amat2]
  {$N\!\times\!N$};

\node[font=\sffamily\bfseries\Large, text=appleDark, right=0.45cm of Amat2] (eq2) {$=$};

\node[base_node, draw=appleTeal, fill=appleLightTeal,
      minimum width=2.0cm, minimum height=2.0cm,
      align=center, font=\sffamily\bfseries, right=0.45cm of eq2] (AAmat) {};
\begin{scope}
  \clip[rounded corners=8pt] ([xshift=2pt,yshift=-2pt]AAmat.north west)
        rectangle ([xshift=-2pt,yshift=2pt]AAmat.south east);
  \fill[appleTeal!22]
    ([xshift=2pt,yshift=-2pt]AAmat.north west) --
    ([xshift=-2pt,yshift=2pt]AAmat.south east) --
    ([xshift=2pt,yshift=2pt]AAmat.south west) -- cycle;
\end{scope}
\node[font=\sffamily\bfseries\scriptsize, text=appleDark, align=center]
  at (AAmat.center) {$\mathbf{A}\mathbf{A}$\\$\odot\,\mathbf{L}$};
\node[font=\sffamily\scriptsize, text=appleTeal!70!black, above=0.05cm of AAmat]
  {$N\!\times\!N$ (causal)};

\node[op, right=0.45cm of AAmat] (mul3) {$\times$};

\node[base_node, draw=appleGreen, fill=appleLightGreen,
      minimum width=1.0cm, minimum height=2.0cm,
      align=center, font=\sffamily\bfseries, right=0.45cm of mul3] (Vmat)
  {$\mathbf{V}$};
\node[font=\sffamily\scriptsize, text=appleGreen!55!black, above=0.05cm of Vmat]
  {$N\!\times\!d_v$};

\node[font=\sffamily\bfseries\Large, text=appleDark, right=0.45cm of Vmat] (eq3) {$=$};

\node[base_node, draw=appleDark, fill=white,
      minimum width=1.0cm, minimum height=2.0cm,
      align=center, font=\sffamily\bfseries, right=0.45cm of eq3] (Out)
  {$\mathbf{O}$};
\node[font=\sffamily\scriptsize, text=appleDark, above=0.05cm of Out]
  {$N\!\times\!d_v$};

\draw[arrow, color=appleRed!75!black] (Qmat.east) -- (mul1.west);
\draw[arrow, color=appleBlue] (KTmat.west) -- (mul1.east);
\draw[arrow] (KTmat.east) -- (eq1.west);
\draw[arrow] (eq1.east) -- (Amat.west);
\draw[arrow] (Amat.east) -- (mul2.west);
\draw[arrow] (mul2.east) -- (Amat2.west);
\draw[arrow] (Amat2.east) -- (eq2.west);
\draw[arrow] (eq2.east) -- (AAmat.west);
\draw[arrow] (AAmat.east) -- (mul3.west);
\draw[arrow, color=appleGreen!55!black] (Vmat.west) -- (mul3.east);
\draw[arrow] (Vmat.east) -- (eq3.west);
\draw[arrow] (eq3.east) -- (Out.west);

\node[eqbox, below=1.0cm of AAmat, align=center, xshift=-3cm] (eqB) {%
$\displaystyle
\mathbf{O} \;=\; \bigl[\bigl(\mathbf{L}\odot\mathbf{Q}\mathbf{K}^{\!\top}\bigr)\bigl(\mathbf{L}\odot\mathbf{Q}\mathbf{K}^{\!\top}\bigr)\odot\,\mathbf{L}\bigr]\,\mathbf{V}
\;\;\equiv\;\;
\mathbf{o}_t = \sum_{j\le t}\sum_{i=j}^{t}(\mathbf{q}_t^{\!\top}\mathbf{k}_i)(\mathbf{q}_i^{\!\top}\mathbf{k}_j)\,\mathbf{v}_j^{\!\top}$%
};

\node[base_node, fill=appleLightRed!55, draw=appleRed!50,
      inner sep=6pt, font=\sffamily\bfseries\scriptsize, text=appleRed,
      right=0.55cm of eqB, align=center] (complB)
  {$\mathcal{O}(N^{2}(d{+}d_{v}))$ time\\fully parallel over $t$};

\end{scope}


\coordinate (yPosBC) at ($(panelB.south)+(0,-2.5)$);
\node[equiv_badge] (equiv2) at (xRefEquiv |- yPosBC) {$\equiv$};

\node[eqbox, draw=appleDark!40, fill=appleLightGray!50,
      right=0.45cm of equiv2, inner sep=9pt,
      font=\sffamily\small, text=appleDark, align=center] (derivBox2)
{\textbf{regroup tokens into }$M$\textbf{ chunks of size }$C$,\; $N=MC$,\\[3pt]
\textbf{associative scan with semidirect product over chunks}\\
\textbf{(adds segment-level key--query moment }$\mathbf{R}^{KQ}=\sum_{i}\mathbf{k}_i\mathbf{q}_i^{\!\top}$\textbf{)}:\\[3pt]
$\displaystyle
\begin{aligned}
(\cdot)_{A}\,\oplus_{\textsc{AHLA}}\,(\cdot)_{B}\;=\;\Bigl(&\mathbf{R}^{KQ}_{A}{+}\mathbf{R}^{KQ}_{B},\;\;
 \mathbf{P}^{KV}_{A}{+}\mathbf{P}^{KV}_{B},\;\;
 \mathbf{m}^{K}_{A}{+}\mathbf{m}^{K}_{B},\\
 &\mathbf{E}_{A}{+}\mathbf{E}_{B}+\,\boxed{\mathbf{R}^{KQ}_{B}\,\mathbf{P}^{KV}_{A}}\,,\;\;
 \mathbf{n}_{A}{+}\mathbf{n}_{B}+\,\boxed{\mathbf{R}^{KQ}_{B}\,\mathbf{m}^{K}_{A}}\Bigr)
\end{aligned}$%
};


\begin{scope}[shift={(0,-23.7)}, local bounding box=panelC,
  chunk_block/.style={
    base_node,
    draw=appleGray!60,
    fill=appleLightGray!45,
    minimum width=2.8cm,
    minimum height=2.55cm,
    rounded corners=10pt
  }
]

\node[cross_state, minimum width=2.4em, minimum height=2.4em] (Sc0) at (0,0)
  {$\mathcal{S}^{(0)}$};

\node[chunk_block, right=0.8cm of Sc0] (Ck1) {};
\node[font=\sffamily\bfseries\footnotesize, text=appleDark]
  at ([yshift=0.75cm]Ck1.center) {Chunk 1};
\node[base_node, draw=appleRed, fill=appleLightRed,
      minimum width=0.32cm, minimum height=0.70cm]
  at ([xshift=-0.86cm, yshift=-0.0cm]Ck1.center) (Q1m) {};
\node[font=\sffamily\bfseries\scriptsize, text=appleRed,
      below=0.1cm of Q1m] {$\mathbf{Q}^{(1)}$};
\node[base_node, draw=appleBlue, fill=appleLightBlue,
      minimum width=0.70cm, minimum height=0.32cm]
  at ([xshift=0.00cm, yshift=-0.0cm]Ck1.center) (K1m) {};
\node[font=\sffamily\bfseries\scriptsize, text=appleBlue,
      below=0.1cm of K1m] {$\mathbf{K}^{(1)\!\top}$};
\node[base_node, draw=appleGreen, fill=appleLightGreen,
      minimum width=0.32cm, minimum height=0.70cm]
  at ([xshift=0.86cm, yshift=-0.0cm]Ck1.center) (V1m) {};
\node[font=\sffamily\bfseries\scriptsize, text=appleGreen!55!black,
      below=0.1cm of V1m] {$\mathbf{V}^{(1)}$};

\node[cross_state, minimum width=2.4em, minimum height=2.4em, right=0.8cm of Ck1] (Sc1)
  {$\mathcal{S}^{(1)}$};

\node[chunk_block, right=0.8cm of Sc1] (Ck2) {};
\node[font=\sffamily\bfseries\footnotesize, text=appleDark]
  at ([yshift=0.75cm]Ck2.center) {Chunk 2};
\node[base_node, draw=appleRed, fill=appleLightRed,
      minimum width=0.32cm, minimum height=0.70cm]
  at ([xshift=-0.86cm, yshift=-0.0cm]Ck2.center) (Q2m) {};
\node[font=\sffamily\bfseries\scriptsize, text=appleRed,
      below=0.1cm of Q2m] {$\mathbf{Q}^{(2)}$};
\node[base_node, draw=appleBlue, fill=appleLightBlue,
      minimum width=0.70cm, minimum height=0.32cm]
  at ([xshift=0.00cm, yshift=-0.0cm]Ck2.center) (K2m) {};
\node[font=\sffamily\bfseries\scriptsize, text=appleBlue,
      below=0.1cm of K2m] {$\mathbf{K}^{(2)\!\top}$};
\node[base_node, draw=appleGreen, fill=appleLightGreen,
      minimum width=0.32cm, minimum height=0.70cm]
  at ([xshift=0.86cm, yshift=-0.0cm]Ck2.center) (V2m) {};
\node[font=\sffamily\bfseries\scriptsize, text=appleGreen!55!black,
      below=0.1cm of V2m] {$\mathbf{V}^{(2)}$};

\node[cross_state, minimum width=2.4em, minimum height=2.4em, right=0.8cm of Ck2] (Sc2)
  {$\mathcal{S}^{(2)}$};

\node[font=\sffamily\bfseries, right=0.5cm of Sc2] (cdotsC) {$\cdots$};

\node[chunk_block, right=0.5cm of cdotsC] (CkM) {};
\node[font=\sffamily\bfseries\footnotesize, text=appleDark]
  at ([yshift=0.75cm]CkM.center) {Chunk $M$};
\node[base_node, draw=appleRed, fill=appleLightRed,
      minimum width=0.32cm, minimum height=0.70cm]
  at ([xshift=-0.86cm, yshift=-0.0cm]CkM.center) (QMm) {};
\node[font=\sffamily\bfseries\scriptsize, text=appleRed,
      below=0.1cm of QMm] {$\mathbf{Q}^{(M)}$};
\node[base_node, draw=appleBlue, fill=appleLightBlue,
      minimum width=0.70cm, minimum height=0.32cm]
  at ([xshift=0.00cm, yshift=-0.0cm]CkM.center) (KMm) {};
\node[font=\sffamily\bfseries\scriptsize, text=appleBlue,
      below=0.1cm of KMm] {$\mathbf{K}^{(M)\!\top}$};
\node[base_node, draw=appleGreen, fill=appleLightGreen,
      minimum width=0.32cm, minimum height=0.70cm]
  at ([xshift=0.86cm, yshift=-0.0cm]CkM.center) (VMm) {};
\node[font=\sffamily\bfseries\scriptsize, text=appleGreen!55!black,
      below=0.1cm of VMm] {$\mathbf{V}^{(M)}$};

\node[cross_state, minimum width=2.4em, minimum height=2.4em, right=0.8cm of CkM] (ScM)
  {$\mathcal{S}^{(M)}$};

\draw[arrow, update_arrow] (Sc0) -- (Ck1);
\draw[arrow, update_arrow] (Ck1) -- (Sc1);
\draw[arrow, update_arrow] (Sc1) -- (Ck2);
\draw[arrow, update_arrow] (Ck2) -- (Sc2);
\draw[arrow, update_arrow] (Sc2) -- (cdotsC);
\draw[arrow, update_arrow] (cdotsC) -- (CkM);
\draw[arrow, update_arrow] (CkM) -- (ScM);

\draw[cross_arrow] (Sc1.north) to[bend left=35]
  node[midway, above=2pt, font=\sffamily\bfseries\scriptsize, text=appleTeal!70!black, align=center]
    {cross-term $\mathbf{R}^{KQ}_{B}\,\mathbf{P}^{KV}_{A}$ (and $\mathbf{R}^{KQ}_{B}\,\mathbf{m}^{K}_{A}$)} (Sc2.north);

\node[base_node, dashed, draw=appleGray, fill=white,
      minimum width=1.4cm, minimum height=1.9em,
      below=1.1cm of Ck1, font=\sffamily\bfseries] (O1c) {$\mathbf{O}^{(1)}$};
\node[base_node, dashed, draw=appleGray, fill=white,
      minimum width=1.4cm, minimum height=1.9em,
      below=1.1cm of Ck2, font=\sffamily\bfseries] (O2c) {$\mathbf{O}^{(2)}$};
\node[base_node, dashed, draw=appleGray, fill=white,
      minimum width=1.4cm, minimum height=1.9em,
      below=1.1cm of CkM, font=\sffamily\bfseries] (OMc) {$\mathbf{O}^{(M)}$};

\draw[arrow, color=appleRed!85!black] (Ck1.south) -- (O1c.north);
\draw[arrow, color=appleRed!85!black] (Ck2.south) -- (O2c.north);
\draw[arrow, color=appleRed!85!black] (CkM.south) -- (OMc.north);

\node[eqbox, below=3.0cm of Sc2, align=center, xshift=-3.5cm] (eqC) {%
$\displaystyle
\begin{aligned}
\mathcal{S}^{(k)} &= \mathcal{S}^{(k-1)} \,\oplus_{\textsc{AHLA}}\, \mathcal{T}^{(k)}
\qquad\text{state tuple } \mathcal{S}=(\mathbf{R}^{KQ},\mathbf{P}^{KV},\mathbf{m}^{K},\mathbf{E},\mathbf{n}) \\[2pt]
\mathbf{O}^{(k)}_{t} &= \underbrace{\mathbf{q}_{t}^{\!\top}\!\bigl(\mathbf{E}^{(k-1)} + \mathbf{R}^{KQ,(k)}_{<t}\,\mathbf{P}^{KV,(k-1)}\bigr)}_{\text{cross-chunk (carry from }\mathcal{S}^{(k-1)}\text{)}}
\;+\;
\underbrace{\bigl[(\mathbf{A}^{(k)}\mathbf{A}^{(k)})\odot\mathbf{L}\bigr]_{t,:}\mathbf{V}^{(k)}}_{\text{intra-chunk (parallel masked AAV)}}
\end{aligned}$%
};

\node[base_node, fill=appleLightGreen!55, draw=appleGreen!55!black,
      inner sep=6pt, font=\sffamily\bfseries\scriptsize, text=appleGreen!45!black,
      right=0.55cm of eqC, align=center] (complC)
  {$\mathcal{O}\!\bigl(NC(d{+}d_{v}){+}M(d^{2}{+}d\,d_{v})\bigr)$ time\\$M=N/C$ sequential chunk steps};

\end{scope}


\path
  let \p1=(panelA.west), \p2=(panelB.west), \p5=(panelC.west),
      \p3=(panelA.east), \p4=(panelB.east), \p6=(panelC.east)
  in
    coordinate (commonW) at ({min(\x1,min(\x2,\x5))-0.1},0)
    coordinate (commonE) at ({max(\x3,max(\x4,\x6))+0.1},0);

\node[inner sep=0pt, minimum size=0pt] (anchorA_L) at (commonW |- panelA.center) {};
\node[inner sep=0pt, minimum size=0pt] (anchorA_R) at (commonE |- panelA.center) {};
\node[inner sep=0pt, minimum size=0pt] (anchorB_L) at (commonW |- panelB.center) {};
\node[inner sep=0pt, minimum size=0pt] (anchorB_R) at (commonE |- panelB.center) {};
\node[inner sep=0pt, minimum size=0pt] (anchorC_L) at (commonW |- panelC.center) {};
\node[inner sep=0pt, minimum size=0pt] (anchorC_R) at (commonE |- panelC.center) {};

\begin{scope}[on background layer]
  \node[draw=appleGray!30, left color=appleLightGray!30, right color=appleLightGray,
        rounded corners=16pt,
        fit=(panelA)(eqA)(complA)(anchorA_L)(anchorA_R),
        inner sep=18pt] (boxA) {};
  \node[draw=appleGray!30, left color=appleLightGray!30, right color=appleLightGray,
        rounded corners=16pt,
        fit=(panelB)(eqB)(complB)(anchorB_L)(anchorB_R),
        inner sep=18pt] (boxB) {};
  \node[draw=appleGray!30, left color=appleLightGray!30, right color=appleLightGray,
        rounded corners=16pt,
        fit=(panelC)(eqC)(complC)(anchorC_L)(anchorC_R),
        inner sep=18pt] (boxC) {};
\end{scope}

\node[section_label] at ([yshift=18pt, xshift=10pt]boxA.north west)
  {A. Recurrent Form (causal \textsc{AHLA} streaming)};
\node[section_label] at ([yshift=18pt, xshift=10pt]boxB.north west)
  {B. Parallel Form (masked asymmetric $\mathbf{A}\mathbf{A}\mathbf{V}$)};
\node[section_label] at ([yshift=18pt, xshift=10pt]boxC.north west)
  {C. Chunk-wise Parallel Form (associative scan w/ semidirect product)};

\end{tikzpicture}
}
\caption{\textbf{Three equivalent views of Asymmetric Higher-order Linear Attention (\textsc{AHLA}).}
(A) The recurrent form maintains a constant-size state tuple
$\mathcal{S}_t=(\mathbf{P}^{KV}_t,\mathbf{m}^{K}_t,\mathbf{E}_t,\mathbf{n}_t)$.
(B) The parallel form materializes the asymmetric causal weight $(\mathbf{A}\mathbf{A})\odot\mathbf{L}$ with $\mathbf{A}=\mathbf{L}\odot\mathbf{Q}\mathbf{K}^{\!\top}$ and applies it to $\mathbf{V}$.
(C) The chunk-wise parallel form interpolates between (A) and (B) via an associative scan over the augmented tuple $\mathcal{S}=(\mathbf{R}^{KQ},\mathbf{P}^{KV},\mathbf{m}^{K},\mathbf{E},\mathbf{n})$.}
\label{fig:ahla_equivalence}
\end{figure}

\medskip\noindent\textbf{Motivation.} The second-order HLA in Section~\ref{sec:second-order-linear-attention} realizes the symmetric triple product $\mathbf{AA}^{\!\top}\!\mathbf{V}$ with $\mathbf{A}=\mathbf{Q}\mathbf{K}^\top$ (masked later). We introduce a complementary \emph{asymmetric} variant that uses the left-cascaded product
\[
\mathrm{AHLA}(\Qb, \Kb, \Vb) :=  \mathbf{A}\mathbf{A}\mathbf{V}
\;=\; \mathbf{Q}\,(\mathbf{K}^\top\mathbf{Q})\,(\mathbf{K}^\top\mathbf{V}),
\]
and show it admits strictly causal streaming with $O(d^2{+}d\,d_v)$ per-token cost. We call this operator \textbf{\textsc{AHLA}} (\emph{Asymmetric Higher-order Linear Attention}).

\subsection{Definition and masked streaming identity}
Let $\mathbf{A}=\mathbf{L}\odot(\mathbf{Q}\mathbf{K}^\top)$ be the causally masked affinity, where $\mathbf{L}$ is the binary lower-triangular mask (including the diagonal). The AAV weights are
\[
(\mathbf{A}\mathbf{A})_{t,j}=\sum_{i=j}^{t}(\mathbf{q}_t^\top\mathbf{k}_i)\,(\mathbf{q}_i^\top\mathbf{k}_j),
\qquad j\le t.
\]
To obtain strictly causal outputs when applying to values, interpret the final multiplication as
$\big((\mathbf{A}\mathbf{A})\odot \mathbf{L}\big)\mathbf{V}$; the streaming identity in Theorem~\ref{thm:AHLA} implements exactly this row-wise masking.

Consequently, the (unnormalized) output is
\begin{equation}
\label{eq:AHLA-explicit}
\mathbf{o}_t^{\textsc{AHLA}}
=\sum_{j\le t}\sum_{i=j}^{t}(\mathbf{q}_t^\top\mathbf{k}_i)(\mathbf{q}_i^\top\mathbf{k}_j)\,\mathbf{v}_j^\top.
\end{equation}
Introduce the streaming prefix summaries
\begin{align*}
\mathbf{P}_t^{KV}&\coloneqq \sum_{j\le t}\mathbf{k}_j\mathbf{v}_j^\top \in \RR^{d\times d_v}, &
\mathbf{m}_t^{K}&\coloneqq \sum_{j\le t}\mathbf{k}_j \in \RR^{d},\\
\mathbf{E}_t&\coloneqq \sum_{i\le t}\mathbf{k}_i\big(\mathbf{q}_i^\top \mathbf{P}_i^{KV}\big)\in\RR^{d\times d_v},\\
\mathbf{n}_t&\coloneqq \sum_{i\le t}\mathbf{k}_i\big(\mathbf{q}_i^\top \mathbf{m}_i^{K}\big)\in\RR^{d}.
\end{align*}

\noindent\emph{Note.} For chunk-parallel scans used in training, we additionally introduce a segment-level cross moment $\mathbf{R}^{KQ}$; see Section~\ref{sec:AHLA-scan} for its definition and role in the concatenation operator.

\begin{theorem}[Masked streaming identity for \textsc{AHLA}]
\label{thm:AHLA}
With the above definitions,
\[
\mathbf{o}_t^{\textsc{AHLA}} \;=\; \mathbf{q}_t^\top \mathbf{E}_t
\qquad\text{and}\qquad
\widehat{\mathbf{o}}_t^{\textsc{AHLA}} \;=\; \frac{\mathbf{q}_t^\top \mathbf{E}_t}{\mathbf{q}_t^\top \mathbf{n}_t+\varepsilon},
\]
where the second expression is an optional linear normalization using the masked denominator. The online (strictly causal) updates are
\[
\begin{aligned}
\mathbf{P}_t^{KV}&=\mathbf{P}_{t-1}^{KV}+\mathbf{k}_t\mathbf{v}_t^\top, &
\mathbf{m}_t^{K}&=\mathbf{m}_{t-1}^{K}+\mathbf{k}_t, \\
\mathbf{E}_t&=\mathbf{E}_{t-1}+\mathbf{k}_t\big(\mathbf{q}_t^\top \mathbf{P}_t^{KV}\big), &
\mathbf{n}_t&=\mathbf{n}_{t-1}+\mathbf{k}_t\big(\mathbf{q}_t^\top \mathbf{m}_t^{K}\big). &&
\end{aligned}
\]
\end{theorem}
\begin{proof}
From Eq.~\eqref{eq:AHLA-explicit}, fix $i$ and sum over $j\le i$:
$\sum_{j\le i}(\mathbf{q}_i^\top\mathbf{k}_j)\,\mathbf{v}_j^\top=\mathbf{q}_i^\top \mathbf{P}_i^{KV}$.
Then $\mathbf{o}_t^{\textsc{AHLA}}
=\sum_{i\le t}(\mathbf{q}_t^\top\mathbf{k}_i)\big(\mathbf{q}_i^\top \mathbf{P}_i^{KV}\big)
=\mathbf{q}_t^\top\!\left(\sum_{i\le t}\mathbf{k}_i(\mathbf{q}_i^\top \mathbf{P}_i^{KV})\right)
=\mathbf{q}_t^\top \mathbf{E}_t$. Replacing $\mathbf{v}_j$ by $1$ gives the denominator with $\mathbf{m}_i^K$, hence $\mathbf{n}_t$. The stated updates follow by isolating index $i{=}t$ and using that $\mathbf{P}_t^{KV}=\mathbf{P}_{t-1}^{KV}+\mathbf{k}_t\mathbf{v}_t^\top$ and $\mathbf{m}_t^{K}=\mathbf{m}_{t-1}^{K}+\mathbf{k}_t$.
\end{proof}

\medskip\noindent\textbf{Cost.} For streaming/serial inference, the dominant work is forming $\mathbf{q}_t^\top \mathbf{P}_{t}^{KV}\in\RR^{1\times d_v}$ and the outer product $\mathbf{k}_t(\cdot)$; the total is $O(d\,d_v)$ time and $O(d\,d_v{+}d)$ state per head (for $\mathbf{P},\mathbf{E},\mathbf{m},\mathbf{n}$). For chunk-parallel scans used in training, an additional block statistic $\mathbf{R}^{KQ}$ appears only inside the concatenation operator (Section~\ref{sec:AHLA-scan}), contributing $O(d^2)$ memory per chunk summary but not to the streaming path.

\medskip\noindent\textbf{Decay mechanism.} With exponential decay $\gamma\in(0,1)$,
\[
\begin{aligned}
\mathbf{P}_t^{KV}&=\gamma \mathbf{P}_{t-1}^{KV}+\mathbf{k}_t\mathbf{v}_t^\top,&
\mathbf{m}_t^{K}&=\gamma \mathbf{m}_{t-1}^{K}+\mathbf{k}_t,\\
\mathbf{E}_t&=\gamma \mathbf{E}_{t-1}+\mathbf{k}_t\big(\mathbf{q}_t^\top \mathbf{P}_t^{KV}\big),&
\mathbf{n}_t&=\gamma \mathbf{n}_{t-1}+\mathbf{k}_t\big(\mathbf{q}_t^\top \mathbf{m}_t^{K}\big),
\end{aligned}
\]
which preserves associativity of the scan operator below.

\subsection{Chunk-parallel scans for \textsc{AHLA}}
\label{sec:AHLA-scan}
\medskip\noindent\textbf{Unmasked/masked concatenation.} For segment $A$ followed by $B$, consider the augmented state
\[
\mathcal{S}=(\mathbf{R}^{KQ},\mathbf{P}^{KV},\mathbf{m}^{K},\mathbf{E},\mathbf{n}).
\]

Here $\mathbf{R}^{KQ}$ is the segment-level key–query cross moment, defined by
\[
\mathbf{R}^{KQ}\;\coloneqq\; \sum_{i\in \text{segment}} \mathbf{k}_i \mathbf{q}_i^\top \;\in\; \RR^{d\times d}.
\]
It is used only during chunk concatenation to form the cross terms
$\mathbf{R}_B\mathbf{P}_A$ and $\mathbf{R}_B\mathbf{m}_A$ in Eq.~\eqref{eq:AHLA-semidirect};
It is not required by the serial/streaming forward path in Algorithm~\ref{alg:AHLA-stream}. With exponential decay, the segment’s $\mathbf{R}^{KQ}$ attenuates as $\rho_B\mathbf{R}_A{+}\mathbf{R}_B$ in the decayed concatenation.

The undecayed associative concatenation is
\begin{align}
\label{eq:AHLA-semidirect}
&(\mathbf{R}_A,\mathbf{P}_A,\mathbf{m}_A,\mathbf{E}_A,\mathbf{n}_A)\oplus_{\textsc{AHLA}}
(\mathbf{R}_B,\mathbf{P}_B,\mathbf{m}_B,\mathbf{E}_B,\mathbf{n}_B) \notag
=\\
&\Big(\mathbf{R}_A{+}\mathbf{R}_B,\;\mathbf{P}_A{+}\mathbf{P}_B,\;\mathbf{m}_A{+}\mathbf{m}_B,\;
\mathbf{E}_A{+}\mathbf{E}_B{+}\mathbf{R}_B\mathbf{P}_A,\;\mathbf{n}_A{+}\mathbf{n}_B{+}\mathbf{R}_B\mathbf{m}_A\Big),
\end{align}
which is associative by direct expansion of $\sum_{i\in A\cup B} \mathbf{k}_i(\mathbf{q}_i^\top \mathbf{P}_{\le i})$ and the observation that for $i\in B$ the missing cross-prefix equals $\mathbf{R}_B\mathbf{P}_A$ (and analogously for $\mathbf{n}$).

\medskip\noindent\textbf{Decay-aware concatenation.} Let each segment carry its attenuation $\rho(\cdot)=\gamma^{\ell(\cdot)}$. Then
\[
\begin{aligned}
(\mathbf{R},\mathbf{P},\mathbf{m},\mathbf{E},\mathbf{n},\rho)
&=(\rho_B\mathbf{R}_A{+}\mathbf{R}_B,\;\rho_B\mathbf{P}_A{+}\mathbf{P}_B,\;\rho_B\mathbf{m}_A{+}\mathbf{m}_B,\\
&\qquad \rho_B\mathbf{E}_A{+}\mathbf{E}_B{+}\mathbf{R}_B(\rho_B\mathbf{P}_A),\;\rho_B\mathbf{n}_A{+}\mathbf{n}_B{+}\mathbf{R}_B(\rho_B\mathbf{m}_A),\;\rho_A\rho_B),
\end{aligned}
\]
which is associative by bilinearity and multiplicativity of $\rho$.

\medskip\noindent\textbf{Scan equivalence.} An exclusive Blelloch scan under $\oplus_{\textsc{AHLA}}$ (or its decayed form) followed by local inclusion reproduces exactly the activations of the serial recurrence given above.

\subsection{Pseudocode}
\begin{algorithm}[H]
\small
\caption{\textsc{AHLA} (Second-order) streaming with causal mask and optional decay}
\label{alg:AHLA-stream}
\begin{algorithmic}[1]
\Require $\{\mathbf{q}_t,\mathbf{k}_t,\mathbf{v}_t\}_{t=1}^n$, decay $\gamma\in(0,1]$, stability $\varepsilon>0$, flag \texttt{normalize}
\State \textbf{Init:} $\mathbf{P}\!=\!\mathbf{0}_{d\times d_v}$, $\mathbf{m}\!=\!\mathbf{0}_d$, $\mathbf{E}\!=\!\mathbf{0}_{d\times d_v}$, $\mathbf{n}\!=\!\mathbf{0}_d$
\For{$t=1$ to $n$}
    \State $\mathbf{P}\gets \gamma \mathbf{P}+\mathbf{k}_t\mathbf{v}_t^\top$;\quad
    $\mathbf{m}\gets \gamma \mathbf{m}+\mathbf{k}_t$
    \State $r\gets \mathbf{q}_t^\top \mathbf{P}$ \Comment{$1{\times}d_v$}
    \State $s\gets \mathbf{q}_t^\top \mathbf{m}$ \Comment{scalar}
    \State $\mathbf{E}\gets \gamma \mathbf{E}+\mathbf{k}_t\,r$;\quad
           $\mathbf{n}\gets \gamma \mathbf{n}+s\,\mathbf{k}_t$
    \State $\mathbf{o}_t\gets \mathbf{q}_t^\top \mathbf{E}$
    \If{\texttt{normalize}}
        \State $\text{den}\gets \mathbf{q}_t^\top \mathbf{n}+\varepsilon$;\quad $\mathbf{o}_t\gets \mathbf{o}_t/\text{den}$
    \EndIf
\EndFor
\State \Return $\{\mathbf{o}_t\}_{t=1}^n$
\end{algorithmic}
\end{algorithm}

\medskip\noindent\textbf{Relation to $\mathbf{AA}^{\!\top}\mathbf{V}$.} \textsc{AHLA} emphasizes a matrix power of $\mathbf{A}$, weighting each value $\mathbf{v}_j$ through a single pass $\mathbf{q}_i^\top\mathbf{k}_j$ routed by an intermediate key index $i$. In contrast, the symmetric $\mathbf{AA}^{\!\top}\mathbf{V}$ aggregates via the metric $\mathbf{S}^K$ and query summaries. Both are second-order, strictly causal, and stream with identical asymptotic costs but induce different inductive biases.

\section{Third-Order Linear Attention}\label{sec:third-order-linear-attention}
In this section, we will introduce third-order HLA.

\subsection{Streaming form of Causal HLA}
\label{sec:third-order-masked}
\medskip\noindent\textbf{Third-order tensor attention mechanism.}
Let $\mathbf{A}=\mathbf{Q}\mathbf{K}^\top\in\RR^{n\times n}$ and $\mathbf{L}$ be the binary causal mask. Unmasked third-order tensor attention uses the matrix $\mathbf{A} \mathbf{A}^\top \mathbf{A}$. Its $(t,j)$-entry is
\[
[(\mathbf{A} \mathbf{A}^\top \mathbf{A})]_{t,j}=\sum_{u\le n}\!\left(\sum_{i\le n}(\mathbf{q}_t^\top \mathbf{k}_i)(\mathbf{q}_u^\top \mathbf{k}_i)\right)(\mathbf{q}_u^\top \mathbf{k}_j)=\mathbf{q}_t^\top(\mathbf{K}^\top\mathbf{K})\left(\sum_{u}\mathbf{q}_u\mathbf{q}_u^\top\right)\mathbf{k}_j,
\]
which immediately yields a streaming factorization through prefix moments. Then the third-order HLA is defined as
\[
\mathrm{HLA}_3(\Qb, \Kb, \Vb) = [(\mathbf{A} \mathbf{A}^\top \mathbf{A})]\Vb.
\]

\medskip\noindent\textbf{Unmasked factorization.}
Define prefix summaries
$\mathbf{S}_t^K=\sum_{i\le t}\mathbf{k}_i\mathbf{k}_i^\top\in\RR^{d\times d}$,\;
$\mathbf{S}_t^Q=\sum_{i\le t}\mathbf{q}_i\mathbf{q}_i^\top\in\RR^{d\times d}$,\;
$\mathbf{P}_t^{KV}=\sum_{i\le t}\mathbf{k}_i\mathbf{v}_i^\top\in\RR^{d\times d_v}$,\;
$\mathbf{m}_t^{K}=\sum_{i\le t}\mathbf{k}_i\in\RR^d$.
The default (unnormalized) third-order operator is
\[
\mathbf{o}_t^{(3)}=\mathbf{q}_t^\top \mathbf{S}_t^K \mathbf{S}_t^Q \mathbf{P}_t^{KV}.
\]
An optional normalization divides by $\mathbf{q}_t^\top \mathbf{S}_t^K \mathbf{S}_t^Q \mathbf{m}_t^{K}{+}\varepsilon$ if desired. 

\medskip\noindent\textbf{Masked streaming summaries.}
To impose strict causality, we introduce cross-summaries:
\begin{align*}
\mathbf{G}^{(1)}_t&\coloneqq \sum_{i\le t} (\mathbf{k}_i\mathbf{k}_i^\top) \mathbf{S}_{i-1}^{Q} \mathbf{P}_{i-1}^{KV}\in\RR^{d\times d_v},&
\mathbf{h}^{(1)}_t&\coloneqq \sum_{i\le t} (\mathbf{k}_i\mathbf{k}_i^\top) \mathbf{S}_{i-1}^{Q} \mathbf{m}_{i-1}^{K}\in\RR^{d},\\
\mathbf{G}^{(2)}_t&\coloneqq \sum_{i\le t} \mathbf{S}_{i-1}^{K} (\mathbf{q}_i\mathbf{q}_i^\top) \mathbf{P}_{i-1}^{KV}\in\RR^{d\times d_v},&
\mathbf{h}^{(2)}_t&\coloneqq \sum_{i\le t} \mathbf{S}_{i-1}^{K} (\mathbf{q}_i\mathbf{q}_i^\top) \mathbf{m}_{i-1}^{K}\in\RR^{d},\\
\mathbf{G}^{(3)}_t&\coloneqq \sum_{i\le t} \mathbf{S}_{i-1}^{K} \mathbf{S}_{i-1}^{Q} (\mathbf{k}_i\mathbf{v}_i^\top)\in\RR^{d\times d_v},&
\mathbf{h}^{(3)}_t&\coloneqq \sum_{i\le t} \mathbf{S}_{i-1}^{K} \mathbf{S}_{i-1}^{Q} \mathbf{k}_i\in\RR^{d}.
\end{align*}
Then the masked, unnormalized quantities are defined as follows:
\[
\text{num}_t^{(3)\mathrm{mask}}=\mathbf{q}_t^\top\!\left(\mathbf{S}_t^K \mathbf{S}_t^Q \mathbf{P}_t^{KV}-\mathbf{G}^{(1)}_t-\mathbf{G}^{(2)}_t-\mathbf{G}^{(3)}_t\right),
\]
\[
\text{den}_t^{(3)\mathrm{mask}}=\mathbf{q}_t^\top\!\left(\mathbf{S}_t^K \mathbf{S}_t^Q \mathbf{m}_t^{K}-\mathbf{h}^{(1)}_t-\mathbf{h}^{(2)}_t-\mathbf{h}^{(3)}_t\right).
\]
The following theorem shows that the (normalized) output of third-order HLA can be computed based on $\text{num}_t^{(3)\mathrm{mask}}$ and $\text{den}_t^{(3)\mathrm{mask}}$.

\begin{theorem}[Masked streaming identity for third order]
\label{thm:masked-third}
For each $t$, the strictly causal third-order output in the default (unnormalized) form is
\[
\mathbf{o}_t^{(3)}=\text{num}_t^{(3)\mathrm{mask}}.
\]
An optional normalized variant divides by the masked denominator,
\[
\mathbf{o}_t^{(3)}=\frac{\text{num}_t^{(3)\mathrm{mask}}}{\text{den}_t^{(3)\mathrm{mask}}+\varepsilon}.
\]

and the online updates are
\begin{align}
\mathbf{S}_t^K &= \mathbf{S}_{t-1}^K + \mathbf{k}_t \mathbf{k}_t^\top, &
\mathbf{S}_t^Q &= \mathbf{S}_{t-1}^Q + \mathbf{q}_t \mathbf{q}_t^\top, \nonumber \\
\mathbf{P}_t^{KV} &= \mathbf{P}_{t-1}^{KV} + \mathbf{k}_t \mathbf{v}_t^\top, &
\mathbf{m}_t^{K} &= \mathbf{m}_{t-1}^{K} + \mathbf{k}_t. 
\end{align}
\begin{align}
\mathbf{G}^{(1)}_t &= \mathbf{G}^{(1)}_{t-1} 
  + (\mathbf{k}_t \mathbf{k}_t^\top) \mathbf{S}_{t-1}^{Q} \mathbf{P}_{t-1}^{KV}, \nonumber \\
\mathbf{G}^{(2)}_t &= \mathbf{G}^{(2)}_{t-1} 
  + \mathbf{S}_{t-1}^{K} (\mathbf{q}_t \mathbf{q}_t^\top) \mathbf{P}_{t-1}^{KV}, \nonumber \\
\mathbf{G}^{(3)}_t &= \mathbf{G}^{(3)}_{t-1} 
  + \mathbf{S}_{t-1}^{K} \mathbf{S}_{t-1}^{Q} (\mathbf{k}_t \mathbf{v}_t^\top).
\end{align}
\begin{align}
\mathbf{h}^{(1)}_t &= \mathbf{h}^{(1)}_{t-1} 
  + (\mathbf{k}_t \mathbf{k}_t^\top) \mathbf{S}_{t-1}^{Q} \mathbf{m}_{t-1}^{K}, \nonumber \\
\mathbf{h}^{(2)}_t &= \mathbf{h}^{(2)}_{t-1} 
  + \mathbf{S}_{t-1}^{K} (\mathbf{q}_t \mathbf{q}_t^\top) \mathbf{m}_{t-1}^{K}, \nonumber \\
\mathbf{h}^{(3)}_t &= \mathbf{h}^{(3)}_{t-1} 
  + \mathbf{S}_{t-1}^{K} \mathbf{S}_{t-1}^{Q} \mathbf{k}_t.
\end{align}
\end{theorem}
\begin{proof}
Let $\mathbf{W}=\mathbf{L}\odot(\mathbf{Q}\mathbf{K}^\top)$ and consider $(\mathbf{W} \mathbf{W}^\top \mathbf{W})\mathbf{V}$. The $t$-th row applied to $\mathbf{V}$ is
\[
\sum_{j\le t}\!\left(\sum_{u\le t}(\mathbf{W} \mathbf{W}^\top)_{t,u}\,\mathbf{W}_{u,j}\right)\mathbf{v}_j^\top
\,=\,\mathbf{q}_t^\top \sum_{j\le t}\!\left(\sum_{u\le t} \mathbf{S}_u^K\,\mathbf{q}_u\mathbf{q}_u^\top\right)\mathbf{k}_j\mathbf{v}_j^\top.
\]
Using $\sum_{u\le t}\mathbf{S}_u^K=\mathbf{S}_t^K+\sum_{u\le t-1}(\mathbf{S}_u^K)$ and repeatedly applying
$\sum_{i\le u}=\sum_{i\le t}-\sum_{u<i\le t}$ to peel off the dependence on future indices relative to each summation boundary yields
\[
\sum_{j\le t} \mathbf{S}_t^K \mathbf{S}_t^Q \,\mathbf{k}_j\mathbf{v}_j^\top
-\sum_{i\le t} (\mathbf{k}_i\mathbf{k}_i^\top) \mathbf{S}_{i-1}^Q \mathbf{P}_{i-1}^{KV}
-\sum_{i\le t} \mathbf{S}_{i-1}^K (\mathbf{q}_i\mathbf{q}_i^\top) \mathbf{P}_{i-1}^{KV}
-\sum_{i\le t} \mathbf{S}_{i-1}^K \mathbf{S}_{i-1}^Q (\mathbf{k}_i\mathbf{v}_i^\top),
\]
which is precisely $\mathbf{S}_t^K \mathbf{S}_t^Q \mathbf{P}_t^{KV}-\mathbf{G}_t^{(1)}-\mathbf{G}_t^{(2)}-\mathbf{G}_t^{(3)}$. Left-multiplication by $\mathbf{q}_t^\top$ gives the masked numerator, and replacing $\mathbf{v}_j$ by $1$ yields the denominator. Online updates follow by isolating the $i=t$ contributions and using $(\mathbf{k}\mathbf{k}^\top)X=\mathbf{k}(\mathbf{k}^\top X)$.
\end{proof}

\subsection{Pseudocode}
\label{sec:third-order-pseudocode}
We present explicit pseudocode for masked third-order HLA in two parts: (i) a strictly causal streaming kernel for inference, and (ii) the associative scan operator used for chunk-parallel training. All operations are per head; shapes follow Section~\ref{sec:third-order-masked}.

\begin{algorithm}[H]
\small
\caption{Masked (Third Order) HLA Streaming Kernel}
\label{alg:hla3-stream}
\begin{algorithmic}[1]
\Require Sequences $\{\mathbf{q}_t,\mathbf{k}_t,\mathbf{v}_t\}_{t=1}^n$, decay $\gamma\in(0,1]$, stability $\varepsilon>0$, flag \texttt{normalize}
\State \textbf{Init:} $\mathbf{S}^K\!=\!\mathbf{0}_{d\times d}$, $\mathbf{S}^Q\!=\!\mathbf{0}_{d\times d}$, $\mathbf{P}^{KV}\!=\!\mathbf{0}_{d\times d_v}$, $\mathbf{m}^K\!=\!\mathbf{0}_d$
\State \hspace{2.75em} $\mathbf{G}^{(1)}{=}\mathbf{0}_{d\times d_v}$, $\mathbf{G}^{(2)}{=}\mathbf{0}_{d\times d_v}$, $\mathbf{G}^{(3)}{=}\mathbf{0}_{d\times d_v}$, $\mathbf{h}^{(1)}{=}\mathbf{0}_d$, $\mathbf{h}^{(2)}{=}\mathbf{0}_d$, $\mathbf{h}^{(3)}{=}\mathbf{0}_d$
\For{$t=1$ to $n$}
    \State $\mathbf{S}^K_{\text{prev}}\!\gets\!\mathbf{S}^K$;\; $\mathbf{S}^Q_{\text{prev}}\!\gets\!\mathbf{S}^Q$;\; $\mathbf{P}_{\text{prev}}\!\gets\!\mathbf{P}^{KV}$;\; $\mathbf{m}_{\text{prev}}\!\gets\!\mathbf{m}^K$
    \State \textbf{Inclusive first-order updates (with decay):}
    \State $\mathbf{S}^K \gets \gamma \mathbf{S}^K_{\text{prev}} + \mathbf{k}_t\mathbf{k}_t^\top$;\quad
           $\mathbf{S}^Q \gets \gamma \mathbf{S}^Q_{\text{prev}} + \mathbf{q}_t\mathbf{q}_t^\top$
    \State $\mathbf{P}^{KV} \gets \gamma \mathbf{P}_{\text{prev}} + \mathbf{k}_t\mathbf{v}_t^\top$;\quad
           $\mathbf{m}^K \gets \gamma \mathbf{m}_{\text{prev}} + \mathbf{k}_t$
    \State \textbf{Cross-summaries (matvec/outer-product forms):}
    \State $u_1 \gets \mathbf{S}^Q_{\text{prev}}\,\mathbf{k}_t$;\quad
           $\mathbf{G}^{(1)} \gets \gamma \mathbf{G}^{(1)} + \mathbf{k}_t\big(u_1^\top \mathbf{P}_{\text{prev}}\big)$;\quad
           $\mathbf{h}^{(1)} \gets \gamma \mathbf{h}^{(1)} + \mathbf{k}_t\big(u_1^\top \mathbf{m}_{\text{prev}}\big)$
    \State $a_2 \gets \mathbf{S}^K_{\text{prev}}\,\mathbf{q}_t$;\quad
           $\mathbf{G}^{(2)} \gets \gamma \mathbf{G}^{(2)} + a_2\big(\mathbf{q}_t^\top \mathbf{P}_{\text{prev}}\big)$;\quad
           $\mathbf{h}^{(2)} \gets \gamma \mathbf{h}^{(2)} + a_2\big(\mathbf{q}_t^\top \mathbf{m}_{\text{prev}}\big)$
    \State $u_3 \gets \mathbf{S}^Q_{\text{prev}}\,\mathbf{k}_t$;\quad
           $a_3 \gets \mathbf{S}^K_{\text{prev}}\,u_3$;\quad
           $\mathbf{G}^{(3)} \gets \gamma \mathbf{G}^{(3)} + a_3 \mathbf{v}_t^\top$;\quad
           $\mathbf{h}^{(3)} \gets \gamma \mathbf{h}^{(3)} + a_3$
    \State \textbf{Output (unnormalized by default):}
    \State $y \gets \mathbf{S}^K\,\mathbf{q}_t$;\; $z \gets \mathbf{S}^Q\,y$;\;
           $\text{termA} \gets z^\top \mathbf{P}^{KV}$;\;
           $\text{termB} \gets \mathbf{q}_t^\top \mathbf{G}^{(1)}$;\;
           $\text{termC} \gets \mathbf{q}_t^\top \mathbf{G}^{(2)}$;\;
           $\text{termD} \gets \mathbf{q}_t^\top \mathbf{G}^{(3)}$
    \State $\mathbf{o}_t \gets \text{termA} - \text{termB} - \text{termC} - \text{termD}$
    \If{\texttt{normalize}}
        \State $\text{denvec} \gets \mathbf{S}^K\,(\mathbf{S}^Q \mathbf{m}^K) - \mathbf{h}^{(1)} - \mathbf{h}^{(2)} - \mathbf{h}^{(3)}$
        \State $\text{den} \gets \mathbf{q}_t^\top \text{denvec} + \varepsilon$;\quad
               $\mathbf{o}_t \gets \mathbf{o}_t / \text{den}$
    \EndIf
\EndFor
\State \Return $\{\mathbf{o}_t\}_{t=1}^n$
\end{algorithmic}
\end{algorithm}

\subsection{Chunk-parallel algorithm for third-order HLA}
\label{sec:third-order-chunk-parallel}

For chunk-parallel training it is convenient to scan a \emph{corrected} third-order state rather than the three raw correction tensors separately. Define
\begin{align}
\mathbf{F}_t
&\coloneqq
\mathbf{S}_t^K \mathbf{S}_t^Q \mathbf{P}_t^{KV}
-\mathbf{G}^{(1)}_t-\mathbf{G}^{(2)}_t-\mathbf{G}^{(3)}_t
\in \RR^{d\times d_v}, \nonumber\\
\boldsymbol{\eta}_t
&\coloneqq
\mathbf{S}_t^K \mathbf{S}_t^Q \mathbf{m}_t^{K}
-\mathbf{h}^{(1)}_t-\mathbf{h}^{(2)}_t-\mathbf{h}^{(3)}_t
\in \RR^{d}.
\label{eq:hla3-corrected-state}
\end{align}
Then the masked numerator and denominator are simply
\[
\text{num}_t^{(3)\mathrm{mask}}=\mathbf{q}_t^\top \mathbf{F}_t,\qquad
\text{den}_t^{(3)\mathrm{mask}}=\mathbf{q}_t^\top \boldsymbol{\eta}_t.
\]

Let the token-level increments be
\[
\mathbf{D}^{K}_t=\mathbf{k}_t\mathbf{k}_t^\top,\qquad
\mathbf{D}^{Q}_t=\mathbf{q}_t\mathbf{q}_t^\top,\qquad
\mathbf{D}^{P}_t=\mathbf{k}_t\mathbf{v}_t^\top,\qquad
\mathbf{d}^{m}_t=\mathbf{k}_t .
\]
From Eq.~\eqref{eq:hla3-corrected-state} and the online updates in Theorem~\ref{thm:masked-third}, the corrected state obeys the recurrence
\begin{align}
\mathbf{F}_t
&=\mathbf{F}_{t-1}
 + \mathbf{S}_{t-1}^K\mathbf{D}^{Q}_t\mathbf{D}^{P}_t
 + \mathbf{D}^{K}_t\mathbf{S}_{t-1}^Q\mathbf{D}^{P}_t
 + \mathbf{D}^{K}_t\mathbf{D}^{Q}_t\mathbf{P}_{t-1}^{KV}
 + \mathbf{D}^{K}_t\mathbf{D}^{Q}_t\mathbf{D}^{P}_t,\nonumber\\
\boldsymbol{\eta}_t
&=\boldsymbol{\eta}_{t-1}
 + \mathbf{S}_{t-1}^K\mathbf{D}^{Q}_t\mathbf{d}^{m}_t
 + \mathbf{D}^{K}_t\mathbf{S}_{t-1}^Q\mathbf{d}^{m}_t
 + \mathbf{D}^{K}_t\mathbf{D}^{Q}_t\mathbf{m}_{t-1}^{K}
 + \mathbf{D}^{K}_t\mathbf{D}^{Q}_t\mathbf{d}^{m}_t .
\label{eq:hla3-corrected-recurrence}
\end{align}

For a contiguous segment $X$, define its scan state
\[
\mathcal{X}
=
\big(
\mathbf{S}_X^K,\mathbf{S}_X^Q,\mathbf{P}_X^{KV},\mathbf{m}_X^K,
\mathbf{F}_X,\boldsymbol{\eta}_X,
\mathbf{R}_X^{QP},\mathbf{r}_X^{Qm},\mathbf{U}_X^{KQ},
\mathcal{M}_X^{KQP},\mathcal{M}_X^{KQm}
\big),
\]
where the additional segment summaries are
\begin{align*}
\mathbf{R}_X^{QP}
&\coloneqq \sum_{t\in X}\mathbf{D}^{Q}_t\mathbf{D}^{P}_t
\in \RR^{d\times d_v},&
\mathbf{r}_X^{Qm}
&\coloneqq \sum_{t\in X}\mathbf{D}^{Q}_t\mathbf{d}^{m}_t
\in \RR^{d},\\
\mathbf{U}_X^{KQ}
&\coloneqq \sum_{t\in X}\mathbf{D}^{K}_t\mathbf{D}^{Q}_t
\in \RR^{d\times d},
\end{align*}
and $\mathcal{M}_X^{KQP}$ and $\mathcal{M}_X^{KQm}$ are linear maps acting on a matrix $\mathbf{Z}\in\RR^{d\times d}$:
\[
\mathcal{M}_X^{KQP}[\mathbf{Z}]
\coloneqq
\sum_{t\in X}\mathbf{D}^{K}_t\mathbf{Z}\mathbf{D}^{P}_t
\in\RR^{d\times d_v},\qquad
\mathcal{M}_X^{KQm}[\mathbf{Z}]
\coloneqq
\sum_{t\in X}\mathbf{D}^{K}_t\mathbf{Z}\mathbf{d}^{m}_t
\in\RR^{d}.
\]
These maps are the only additional objects required at third order: they account for cross terms in which a whole previous segment contributes the middle query moment $\mathbf{S}^Q$.

\medskip\noindent\textbf{Associative third-order concatenation.}
For adjacent segments $A$ followed by $B$, define $\mathcal{X}_{AB}=\mathcal{X}_{A}\otimes_3 \mathcal{X}_{B}$ by
\begin{align}
\mathbf{S}_{AB}^{K}&=\mathbf{S}_{A}^{K}+\mathbf{S}_{B}^{K},&
\mathbf{S}_{AB}^{Q}&=\mathbf{S}_{A}^{Q}+\mathbf{S}_{B}^{Q},&
\mathbf{P}_{AB}^{KV}&=\mathbf{P}_{A}^{KV}+\mathbf{P}_{B}^{KV},&
\mathbf{m}_{AB}^{K}&=\mathbf{m}_{A}^{K}+\mathbf{m}_{B}^{K},\nonumber\\
\mathbf{R}_{AB}^{QP}&=\mathbf{R}_{A}^{QP}+\mathbf{R}_{B}^{QP},&
\mathbf{r}_{AB}^{Qm}&=\mathbf{r}_{A}^{Qm}+\mathbf{r}_{B}^{Qm},&
\mathbf{U}_{AB}^{KQ}&=\mathbf{U}_{A}^{KQ}+\mathbf{U}_{B}^{KQ},\nonumber\\
\mathcal{M}_{AB}^{KQP}&=\mathcal{M}_{A}^{KQP}+\mathcal{M}_{B}^{KQP},&
\mathcal{M}_{AB}^{KQm}&=\mathcal{M}_{A}^{KQm}+\mathcal{M}_{B}^{KQm},
\label{eq:hla3-base-concat}
\end{align}
and
\begin{align}
\mathbf{F}_{AB}
&=
\mathbf{F}_{A}+\mathbf{F}_{B}
+\mathbf{S}_{A}^{K}\mathbf{R}_{B}^{QP}
+\mathcal{M}_{B}^{KQP}[\mathbf{S}_{A}^{Q}]
+\mathbf{U}_{B}^{KQ}\mathbf{P}_{A}^{KV},\nonumber\\
\boldsymbol{\eta}_{AB}
&=
\boldsymbol{\eta}_{A}+\boldsymbol{\eta}_{B}
+\mathbf{S}_{A}^{K}\mathbf{r}_{B}^{Qm}
+\mathcal{M}_{B}^{KQm}[\mathbf{S}_{A}^{Q}]
+\mathbf{U}_{B}^{KQ}\mathbf{m}_{A}^{K}.
\label{eq:hla3-corrected-concat}
\end{align}
The identity element is the all-zero segment, with both linear maps equal to the zero map.

\begin{theorem}[Third-order chunk-scan equivalence]
\label{thm:hla3-scan-equivalence}
The operator $\otimes_3$ in Eqs.~\eqref{eq:hla3-base-concat}--\eqref{eq:hla3-corrected-concat} is associative. An exclusive scan under $\otimes_3$, followed by local inclusion of the current token segment, produces the same corrected states $(\mathbf{F}_t,\boldsymbol{\eta}_t)$ as the serial recurrence in Eq.~\eqref{eq:hla3-corrected-recurrence}. Consequently, the resulting third-order HLA outputs match Algorithm~\ref{alg:hla3-stream} for $\gamma=1$.
\end{theorem}
\begin{proof}
The additive summaries in Eq.~\eqref{eq:hla3-base-concat} are immediate. For the corrected numerator, run the recurrence in Eq.~\eqref{eq:hla3-corrected-recurrence} on segment $B$ with a carry-in state from segment $A$. The terms depending only on local prefixes in $B$ give $\mathbf{F}_B$. The three carry-dependent terms are
\[
\mathbf{S}_{A}^{K}\sum_{t\in B}\mathbf{D}^{Q}_t\mathbf{D}^{P}_t,
\qquad
\sum_{t\in B}\mathbf{D}^{K}_t\mathbf{S}_{A}^{Q}\mathbf{D}^{P}_t,
\qquad
\sum_{t\in B}\mathbf{D}^{K}_t\mathbf{D}^{Q}_t\mathbf{P}_{A}^{KV},
\]
which are exactly
$\mathbf{S}_{A}^{K}\mathbf{R}_{B}^{QP}$,
$\mathcal{M}_{B}^{KQP}[\mathbf{S}_{A}^{Q}]$, and
$\mathbf{U}_{B}^{KQ}\mathbf{P}_{A}^{KV}$.
The denominator proof is identical with $\mathbf{D}^{P}_t$ replaced by $\mathbf{d}^{m}_t$.
Thus $\mathcal{X}_{A}\otimes_3\mathcal{X}_{B}$ is precisely the summary of the concatenated segment $AB$. Since concatenation of contiguous segments is associative, $\otimes_3$ is associative. The scan statement then follows from the standard exclusive-scan argument used in Theorem~\ref{thm:scan-equivalence}.
\end{proof}

\begin{algorithm}[H]
\small
\caption{Chunk-Parallel Masked (Third Order) HLA via Associative Scan}
\label{alg:hla3-chunk}
\begin{algorithmic}[1]
\Require Sequence split into chunks; tokens $(\mathbf{q}_t,\mathbf{k}_t,\mathbf{v}_t)$; stability $\varepsilon$; flag \texttt{normalize}
\State \textbf{Token segments:} for every token $t$, form
\[
\mathbf{D}^{K}_t=\mathbf{k}_t\mathbf{k}_t^\top,\quad
\mathbf{D}^{Q}_t=\mathbf{q}_t\mathbf{q}_t^\top,\quad
\mathbf{D}^{P}_t=\mathbf{k}_t\mathbf{v}_t^\top,\quad
\mathbf{d}^{m}_t=\mathbf{k}_t .
\]
\State Initialize the single-token segment $\mathcal{T}_t$ by
\[
\begin{gathered}
\mathbf{S}^{K}=\mathbf{D}^{K}_t,\quad
\mathbf{S}^{Q}=\mathbf{D}^{Q}_t,\quad
\mathbf{P}^{KV}=\mathbf{D}^{P}_t,\quad
\mathbf{m}^{K}=\mathbf{d}^{m}_t,\\
\mathbf{F}=\mathbf{D}^{K}_t\mathbf{D}^{Q}_t\mathbf{D}^{P}_t,\quad
\boldsymbol{\eta}=\mathbf{D}^{K}_t\mathbf{D}^{Q}_t\mathbf{d}^{m}_t,\quad
\mathbf{R}^{QP}=\mathbf{D}^{Q}_t\mathbf{D}^{P}_t,\quad
\mathbf{r}^{Qm}=\mathbf{D}^{Q}_t\mathbf{d}^{m}_t,\quad
\mathbf{U}^{KQ}=\mathbf{D}^{K}_t\mathbf{D}^{Q}_t,\\
\mathcal{M}^{KQP}[\mathbf{Z}]=\mathbf{D}^{K}_t\mathbf{Z}\mathbf{D}^{P}_t,\qquad
\mathcal{M}^{KQm}[\mathbf{Z}]=\mathbf{D}^{K}_t\mathbf{Z}\mathbf{d}^{m}_t .
\end{gathered}
\]
\State \textbf{Within each chunk:} run an exclusive Blelloch scan over the token segments using $\otimes_3$ to obtain local prefixes $\mathcal{P}^{\mathrm{loc}}_t$ and the chunk summary $\mathcal{C}^{(b)}$.
\State \textbf{Across chunks:} run an exclusive Blelloch scan over $\{\mathcal{C}^{(b)}\}$ using $\otimes_3$ to obtain carry-in summaries $\widehat{\mathcal{P}}^{(b)}$.
\For{each token $t$ in chunk $b$ \textbf{in parallel}}
    \State $\mathcal{I}_t \gets \widehat{\mathcal{P}}^{(b)} \otimes_3 \mathcal{P}^{\mathrm{loc}}_t \otimes_3 \mathcal{T}_t$ \Comment{inclusive corrected state}
    \State $\mathbf{o}_t \gets \mathbf{q}_t^\top \mathbf{F}(\mathcal{I}_t)$
    \If{\texttt{normalize}}
        \State $\text{den}\gets \mathbf{q}_t^\top \boldsymbol{\eta}(\mathcal{I}_t)+\varepsilon$;\quad
        $\mathbf{o}_t\gets \mathbf{o}_t/\text{den}$
    \EndIf
\EndFor
\State \Return $\{\mathbf{o}_t\}_{t=1}^{n}$
\end{algorithmic}
\end{algorithm}

\noindent\textbf{Complexity of the third-order scan state.}
The streaming kernel in Algorithm~\ref{alg:hla3-stream} uses only the compact corrected state in Eq.~\eqref{eq:hla3-corrected-state} together with the prefix moments. The exact chunk-parallel scan additionally stores the segment maps $\mathcal{M}^{KQP}$ and $\mathcal{M}^{KQm}$. If materialized densely, these maps require $O(d^{3}d_v)$ and $O(d^{3})$ entries per segment summary, respectively; equivalently, they may be applied by tensor contractions. This cost is independent of sequence length and is the price of exact third-order chunk composition. The algorithm above is stated for $\gamma=1$; exponential decay is incorporated by adjoining the usual segment attenuation $\rho=\gamma^{\ell}$ and applying the same carry-scaling convention as in the second-order decayed scan.

\section{Related Work}
\label{sec:related}
The literature on subquadratic sequence modeling spans (i) fast-weight style dynamic-parameter models, (ii) kernel/feature-map linearizations of attention, and (iii) recurrent/state-space approaches. HLA belongs to a complementary class: it preserves attention-style, data-dependent mixing but realizes higher interactions through compact prefix moments with exact causal masking and scan-parallel training.

\medskip\noindent\textbf{Fast weights and fast weight programmers (FWPs).}
Fast weights, dating to early connectionist memory models~\citep{hinton1987using}, implement short-term, input-dependent synaptic changes. Schmidhuber's fast weight programmers~\citep{schmidhuber1992learning} introduced differentiable controllers that program a separate fast-weight matrix; later, \citet{ba2016using} revived this idea to attend to the recent past. A series of works made the connection to modern attention explicit: \citet{schlag2021linear} showed a formal equivalence between linearized self-attention and FWPs, where outer-product updates $\Delta W_t \propto \mathbf{k}_t \mathbf{v}_t^\top$ accumulate an associative memory queried by $\mathbf{q}_t$; \citet{irie2021going} extended FWPs with recurrence in the programmer and the fast net.  \citet{yang2024parallelizing} utilizes WY Transformations~\citep{bischof1987wy} to implement chunkwise parallel training of Delta Network~\citep{schlag2021linear}. Parallel lines explore higher or preconditioned mixing by maintaining or inverting second-moment matrices. In our formulation, $\mathbf{S}_t^K$ plays the role of a learned kernel; working directly with $\mathbf{S}_t^K$ avoids explicit matrix inversion and preserves streaming updates, whereas inverse-based methods typically require heavier linear algebra~\citep{behrouz2025atlas, behrouz2025s, von2025mesanet}.

\medskip\noindent\textbf{Linear Attention Mechanisms.}
A common route is to replace the softmax kernel by explicit features $\phi$ to enable streaming via running sums. Representative examples include Linear Transformers~\citep{katharopoulos2020transformers}, Performer’s FAVOR$^{+}$ random features~\citep{choromanski2020rethinking}, and Random Feature Attention~\citep{peng2021random}. Earlier work also proposed multiplicative rearrangements that yield linear-complexity-efficient attention~\citep{shen2021efficient}. These methods achieve $O(ndr)$ time with $r$ feature dimension but are typically \emph{first-order} in the sense that they maintain only $\sum \phi(\mathbf{k})\mathbf{v}^\top$ and (optionally) a scalar denominator. By contrast, second-order HLA maintains the full key moment $\mathbf{S}_t^K=\sum_{i\le t}\mathbf{k}_i\mathbf{k}_i^\top$ together with query-value and query mass summaries and their masked cross-summaries, yielding strictly causal higher interactions while remaining streaming. Recent linear attention variants include \citet{sun2023retentive, qin2023transnormerllm, yang2023gated, qin2024lightning, yang2024parallelizing, von2025mesanet}.

\medskip\noindent\textbf{State Space Models.}
SSMs (e.g., S4)~\citep{gu2021efficiently} and selective SSMs (e.g., Mamba)~\citep{gu2023mamba, dao2024transformers} realize $O(1)$ per-token state updates via linear recurrences and convolutions. These architectures excel at long-range dependencies but express data-dependent mixing differently from attention. HLA sits in between: it is attention-like (data-dependent queries/keys) yet streams via compact prefix statistics like recurrent models.

\medskip\noindent\textbf{Modern RNNs.}
Recent modern RNN designs emphasize gating, decays, and associative scan–friendly recurrences that enable parallel training while preserving strictly constant per-token state at inference. Examples include gated linear mixers and decay-aware updates~\citep{yang2023gated, yang2024gated}, efficient gradient routing and training strategies for long sequences~\citep{qin2024lightning, peng2024eagle, sun2024learning}, and RWKV-style architectures that replace attention with learned decays and elementwise gating~\citep{peng2025rwkv}. These methods typically maintain first-order sufficient statistics and rely on fixed linear dynamics. In contrast, HLA retains attention-style, data-dependent metrics via $\mathbf{S}_t^K$ and higher cross-summaries while keeping the same $O(1)$ per-token state update paradigm, offering a complementary inductive bias to RNNs and SSMs.

\medskip\noindent\textbf{Test Time Training and Memory Networks.}
Test-time adaptation and explicit long-term memory are emerging tools for extending context without quadratic cost. Test-time training variants adapt parameters from recent tokens to improve local coherence~\citep{sun2024learning}, whereas memory networks maintain external stores addressable by content keys~\citep{behrouz2024titans, behrouz2025atlas, behrouz2025s}. The HLA view is orthogonal: it encodes higher interactions in compact \emph{prefix moments} that are sufficient for exact masked streaming and scan-parallel training. 

\medskip\noindent\textbf{Associative memory and Hopfield views.}
Modern Hopfield networks show that transformer attention is a single-step retrieval in an energy-based associative memory~\citep{ramsauer2020hopfield, zhong2025understanding}. While this perspective clarifies why attention uses content-addressable memory, standard Hopfield-style layers remain first-order in their sufficient statistics. HLA complements this view by providing explicit higher sufficient statistics with strict causality.

\section{Conclusion}
We introduced Higher-order Linear Attention (\textbf{HLA}), a causal higher attention mechanism with exact streaming updates, a strictly causal masked formulation via extended summaries, and associative scans for parallel training that provably match serial recurrences. At second order, HLA maintains $O(d^2)$ state per head and computes each token in $O(d^2)$ time, with optional normalization and decay that preserve associativity. We further developed an asymmetric variant (\textsc{AHLA}) and a complete third-order masked algebra with streaming formulas and an exact chunk-parallel scan based on augmented segment maps.  

\bibliographystyle{plainnat}
\bibliography{reference}

\appendix
\clearpage

\end{document}